\documentclass{article}

    \usepackage[final]{neurips_2022}

\usepackage[utf8]{inputenc} 
\usepackage[T1]{fontenc}    

\usepackage{hyperref}       
\usepackage{url}            
\usepackage{times}
\usepackage{ulem}
\usepackage{latexsym}
\usepackage{microtype}
\usepackage{graphicx}
\usepackage[scriptsize]{subfigure}
\usepackage{caption}
\usepackage{amsmath,amsfonts,amssymb}
\usepackage{bm}
\usepackage{nicefrac}
\usepackage{microtype}
\usepackage{algorithm}
\usepackage{algorithmic}
\usepackage{booktabs} 
\usepackage[export]{adjustbox}

\usepackage{color,xcolor}
\usepackage{epsfig}
\usepackage{graphicx}

\usepackage{adjustbox}
\usepackage{tabularx}
\usepackage{array}
\usepackage{booktabs} 
\usepackage{colortbl}
\usepackage{float,wrapfig}
\usepackage{hhline}
\usepackage{multirow}
\usepackage{todonotes}

\definecolor{MyDarkBlue}{rgb}{0,0.08,1}
\definecolor{MyDarkGreen}{rgb}{0.02,0.6,0.02}
\definecolor{MyDarkRed}{rgb}{0.8,0.02,0.02}
\definecolor{MyDarkOrange}{rgb}{0.80,0.4,0.04}
\definecolor{MyPurple}{RGB}{111,0,255}
\definecolor{MyRed}{rgb}{1.0,0.0,0.0}
\definecolor{MyGold}{rgb}{0.75,0.6,0.12}
\definecolor{MyDarkgray}{rgb}{0.66, 0.66, 0.66}

\newcommand{\sidi}[1]{}
\newcommand{\sidistoryline}[1]{}
\newcommand{\tao}[1]{}
\newcommand{\violet}[1]{}
\newcommand{\violettodo}[1]{}

\newcommand{\sequence}{\boldsymbol{s}}
\newcommand{\known}{\boldsymbol{s}^{o}}

\newcommand{\token}{x}

\newcommand{\dual}[1]{\multirow{2}{*}{#1}}
\newcommand{\position}{\boldsymbol{l}}
\renewcommand{\emph}[1]{\textit{#1}}
\newcommand{\tinysection}[1]{\textbf{#1}~~}
\newcommand{\mat}[1]{\mathbf{#1}}

\newcommand{\model}{\textsc{InsNet}}
\newcommand{\papertitle}{\model: An Efficient, Flexible, and Performant Insertion-based Text Generation Model} 

\title{\papertitle}

\author{Sidi Lu,~~Tao Meng,~~Nanyun Peng\\
  University of California, Los Angeles\\
  \texttt{\{sidilu, tmeng, violetpeng\}@cs.ucla.edu} }

\begin{document}

\maketitle

\begin{abstract}
\violet{summarize the highlights of InsNet: efficient training, flexible decoding, expressive)}
We propose \model{}, an expressive insertion-based text generator with efficient training and flexible decoding (parallel or sequential). Unlike most existing insertion-based text generation works that require re-encoding of the context after each insertion operation and thus are inefficient to train, \model{} only requires one pass of context encoding for the entire sequence during training by introducing a novel insertion-oriented position encoding and a light-weighted slot representation strategy to enable computation sharing.   
\violet{explain flexible decoding} Furthermore, we propose an algorithm \model{}-Dinic to better determine the parallelization of insertion operations that provides a controllable switch between parallel and sequential decoding, making it flexible to handle more parallelizable tasks such as machine translation with efficient decoding, or less parallelizable tasks such as open-domain text generation to guarantee high-quality outputs. \violet{we show strong empirical results.}
Experiments on two lexically constrained text generation datasets and three machine translation datasets demonstrate \model's advantages over previous insertion-based methods in terms of training speed, inference efficiency, and generation quality.\footnote{The code will be publicly released soon on github.}

\end{abstract}

\section{Introduction}

\violet{why insertion-based generation is important} Insertion-based text generation that formulates the generation process as a sequence of token insertion operations has received increasing attention in recent years. 
There are two major advantages of insertion-based generation over the prevalent left-to-right auto-regressive paradigm: 1) It reduces the decoding cost to sub-linear \emph{w.r.t.} the sequence length with parallel decoding \citep{stern2019insertion,gu2019levenshtein}, and 2) the flexible insertion orders may better recover/utilize the underlying linguistic structures of languages~\citep{welleck2019non,gu2019insertion}. 

\begin{figure} 
\begin{minipage}[h]{.45\linewidth} 
\centering 
\includegraphics[width=\columnwidth]{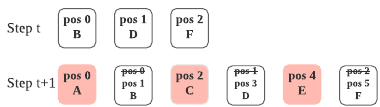}
\caption{The absolute position information of each token is volatile for insertion-based models. Thus, naive models with absolute position encoding have to re-encode the sequence after each insertion operation during training.} 
  \label{fig:volatility_pos}
\end{minipage} 
\hspace{20pt}
\begin{minipage}[h]{.45\linewidth} 
\centering 
\resizebox{1.0\linewidth}{!}{
  \begin{tabular}{l@{ }cc@{}c@{}}
    \toprule
    \dual{Model} & \#Re-Enc. & \#Re-Enc. & \dual{PosInfo} \\ 
    &  w/ ParaDec &  w/ SeqDec \\
   \midrule 
   Ins. Trans. & $\Theta(\log n)$ & \textcolor{MyDarkRed}{$O(n)$} & Absolute \\
   Lev. Trans. & $\Theta(\log n)$ & \textcolor{MyDarkRed}{$\mathbf{\Omega(n)}$} & Absolute \\
   NMSTG & \textcolor{MyDarkRed}{$O(n)$} & \textcolor{MyDarkRed}{$O(n)$} & \textcolor{MyDarkRed}{Markovian}/Absolute \\ 
   InDIGO & \textcolor{MyDarkRed}{N/A} & $O(1)$  & \textcolor{MyDarkRed}{Direction-only} \\
   \midrule
   \model{} (Ours) & $O(1)$ & $O(1)$  & Relative \\
  \bottomrule
  \end{tabular}
 }
\captionof{table}{Comparisons between \model{} and existing insertion-based models regarding training-time re-encoding steps (\emph{\#Re-Enc.} columns) and how the models encode positional information (the \emph{PosInfo} column).
} 
\label{tab:comparative_benefits}
\end{minipage} 
\end{figure}

\violet{major challenge of ins-based generation: training efficiency. two imperfect solutions: parallel decoding and simplified model which lose some model capability.} 
However, this new paradigm of text generation brings unique challenges, mostly in the training efficiency. Unlike left-to-right auto-regressive decoders which monotonically expand the context, the insertion operations complicate the position information of each token as the context expands. Concretely, as is shown in Figure~\ref{fig:volatility_pos}, the absolute position of a token in a sequence constantly changes along with the insertion operations.
As a result, a naive implementation of insertion-based models (e.g., ~\cite{stern2019insertion,gu2019levenshtein}) needs to re-encode the context with updated positional information for each token as the insertions proceed, yielding inefficient training with $O(n)$ times of context re-encoding (with $n$ indicating the sequence length).  

\sidistoryline{current solutions towards the aforementioned challenges and their limitation} \sidistoryline{ins-based models w/ parallel decoding} 
To tackle this problem, previous insertion-based generation models such as Insertion Transformer (InsT)~\citep{stern2019insertion} and Levenshtein Transformer (LevT) \citep{gu2019levenshtein} propose parallel token insertion to reduce the insertion/re-encoding steps from $O(n)$ to $\Theta(\log n)$ for both training and inference. However, while it works well for machine translation, such parallel insertion falls short on high-entropy generation tasks such as open-domain dialogue systems\citep{li2017adversarial}, creative generation such as stories \citep{yao2019plan,goldfarb2020content,han2022go}, poetry \citep{manurung2000poetry,tian2022sonnet}, and humor generation \citep{hempelmann2008humor,he2019pun,Mittal2022ambipun,tian2022unified,sun2022context}, where the output is open-ended and the tokens are usually highly dependent on each other, thus cannot be inserted in parallel. 
\sidistoryline{sequentially-decoded ins-based models} 
Another thread of work seeks to maintain sequential insertions but simplify the position encoding for the tokens so that they do not change as the insertion operations proceed~\citep{gu2019insertion} to save computation. However, this sacrifices model capability and limits its applications to complex sequences.

\sidistoryline{Bullet points of our contribution, a brief explanation of what we did in InsNet: 1) enable computation sharing without compromising model capacity via designing and incorporating an insertion-oriented, relative positional encoding called "offset". Previous encodings in"Offset" can remain unchanged as the insertion-based generation proceeds and in each step offset accurately record the updated spatial relation in even the pair-wise level. 3) slot representation from the aggregation module allows fast position/token likelihood prediction for both sequential/parallel decoding 4) On top of this good representation, we propose InsNet-Dinic to parallelize the insertions with subject to minimized sequence likelihood discrepancy between sequential/parallel decoding under a unified framework} 
In this paper, we propose \model{}, which addresses the training efficiency issue of insertion-based text generators by enabling computation sharing similar as that in vanilla decoder transformers. 
\textbf{Our first contribution} is an insertion-oriented, relative positional encoding coined \textit{offset} that allows \model{} to achieve computation sharing without compromising model capacity. 
At each insertion step, previously computed position encodings of the existing tokens remain unchanged, while the position encodings of newly inserted tokens accurately recording the updated pairwise token spatial relations of all the inserted tokens.
A corresponding offset can be efficiently computed for any given insertion order with a novel process named \textit{offset compression}. 
In this way, we avoid expensive re-encoding of the context, and based on the encoded context, \textbf{our second contribution} is the design of an effective aggregation strategy that allows us to parallelly generate expressive slot representations for every slot in the partially generated sequence, flexibly support both sequential and parallel decoding. 
\textbf{Finally}, 
inspired by the layerization idea of Dinic's algorithm\citep{dinitz2006dinitz}, we propose an algorithm for \model{} to better determine the parallelization of insertion operations in order to reduce the likelihood discrepancy before and after the parallelization.
With all the components, \model{} is a novel framework that enables efficient training, flexible decoding, and expressive positional encoding. Table~\ref{tab:comparative_benefits} shows a comparison between \model{} and all prior insertion-based models.


\section{Related Works and Background}
\sidistoryline{brief intro to AR models. make our audience aware that AR models assume left-to-right generation order} \tinysection{Auto-regressive Language Models} minimize the negative-log-likelihood of a sequence of n tokens $\sequence_{<n} = [\token_0, \token_1, ..., \token_{n-1}]$ 
with a left-to-right factorization. With the transformer architecture \citep{vaswani2017attention}, each step of likelihood estimation can be calculated in parallel while sharing the prefix context encoding calculations. This makes it possible to build powerful and efficient text generation models, like the GPT family \citep{radford2018gpt1,radford2019gpt2,brown2020gpt3}. A lot of successful applications are based on this paradigm of models, such as automatic story generation~\citep{yao2019plan,tan2020progressive}, image captioning~\citep{vinyals2015show,pmlr-v37-xuc15}, machine translation~\citep{dzmitry2015neural,liu2020multilingual}, and dialogue system~\citep{li2017end,li2017adversarial}.

\sidistoryline{detailed explanation of InsT and LevT} \tinysection{Insertion-based Models with Parallel Decoding} Insertion transformer~\citep{stern2019insertion} (InsT) proposes a 
design for insertion-based text generation. In each step, a bi-directional encoder transformer is performed on the partial sequence to compute the representation for each candidate slot between every two consecutive positions. Then, a model for the joint distribution of position-token is built to insert one or more token(s). 
%
%
Variants of InsT share the common atomic objective that models the \emph{step log-likelihood}. On step $t$ where a token $\token_{i \downarrow i+1}$ is inserted in slot $\position_{i\downarrow i+1}$ between position $i$ and $i + 1$ of context $\known_t$, the \emph{step log-likelihood} can be written as: 
  
\begin{equation*}
    \mathop{\log} p(\token_{i\downarrow i+1},\position_{i\downarrow i+1}|\known_t) = \mathop{\log} p_{position}(i+1|\known_t) + \mathop{\log} p_{token}(\token_{i \downarrow i+1}| e(\known_t)_i \oplus e(\known_t)_{i+1}),
\end{equation*}
where $e(\cdot)_i$ stands for the i-th position of bi-directional encoding of the sequence and $\oplus$ stands for vector concatenation. InsT adopts the original absolute positional encoding of transformers, the representation of the generated sequence has to be \emph{completely re-encoded} after each step of context expansion to match the position changes of tokens. The expectation of the negative log step likelihood over all permitted context-insertion pairs at each step is computed as the \emph{step loss}. The step losses from the first step to the last one are summed up as the \emph{sequence loss}. Benefiting from the partially parallelized prediction of tokens, InsT can reduce the number of re-encoding steps to $\Theta(\log n)$.

Levenshtein Transformer \citep{gu2019levenshtein} (LevT) contains two phases during generation: 1) Insertion phase: it first uses a similar strategy as InsT, but only inserts placeholders; an MLM is applied to fulfill the placeholders. 2) Deletion phase: the model is trained as a token-wise discriminator to determine where to delete, under the evaluation of the Levenshtein distance by dynamic programming. In practice, this would result in a slower process compared to InsT, but better generation quality as it alleviates the incoherence caused by parallel insertions.

\sidistoryline{discussion about the difficulty in implementing sequentially-decoded insertion-based models. Brief intro to NMSTG, InDIGO and their detour. Also state that it's not trivial to support parallel decoding in InDIGO as it does not have slot-level representation} \tinysection{Sequential Insertion-based Model} Previous exploration in insertion-based models mostly assumed the usage of parallel decoding in each decoding step, resulting in a partially auto-regressive procedure. For those are trained to only generate one token per decoding step, a vanilla implementation would result in an $O(n)$ factor in training time complexity. NMSTG (Non-monotonic Sequential Text Generation, \citet{welleck2019non}) is one of the first attempts at modeling a non-monotonic sequential insertion-based generation process. It constrains the dependencies of each inserted token to pseudo-Markovian on an expansion tree. InDIGO\citep{gu2019insertion} is proposed as a sequentially-decoded insertion-based model with the transformer architecture. It supports the efficient likelihood estimation by working around the aforementioned \emph{volatility problem} (see Figure~\ref{fig:volatility_pos}) at a cost of omitting the distance information in its position encoding, and is thus able to adopt the conventional computation sharing trick to boost the multiplicative factor in training time complexity for each sequence to as fast as $O(1)$. However, InDIGO uses a encoding of all previously inserted tokens to predict the next token regardless of their tentative position. It is only after the next token is predicted, the inserted position for it is predicted. Thus, there's no trivial solution to use InDIGO as a (partially) non-autoregressive insertion-based generator.

\sidistoryline{mentioning the legacies of Transformer-XL/XLNet} \tinysection{Absolute vs. Relative Position Embedding} The original transformer \citep{vaswani2017attention} uses sinusoidal, absolute position embeddings. Relative position embedding in transformers \citep{shaw2018self,dai-etal-2019-transformerxl} was originally proposed to make the modeling of spatial relation invariant to position translation, and to improve the long-term dependency performance of the model. 
In replacement of absolute positions, which are tied to each token in the sequence, relative positions try to encode the spatial layout of the tokens with a matrix that records a directed distance from the column token to the row token. We further develop the idea of relative position embedding as the key component to overcome the issue of volatile position information of absolute positions in an insertion-based generation process. A noticeable fact is that InDIGO's implementation of position encoding can also be regarded as a direction-only relative position embedding system.

\begin{figure}[t]
\vspace{-15pt}
  \centering
  \includegraphics[width=0.99\columnwidth]{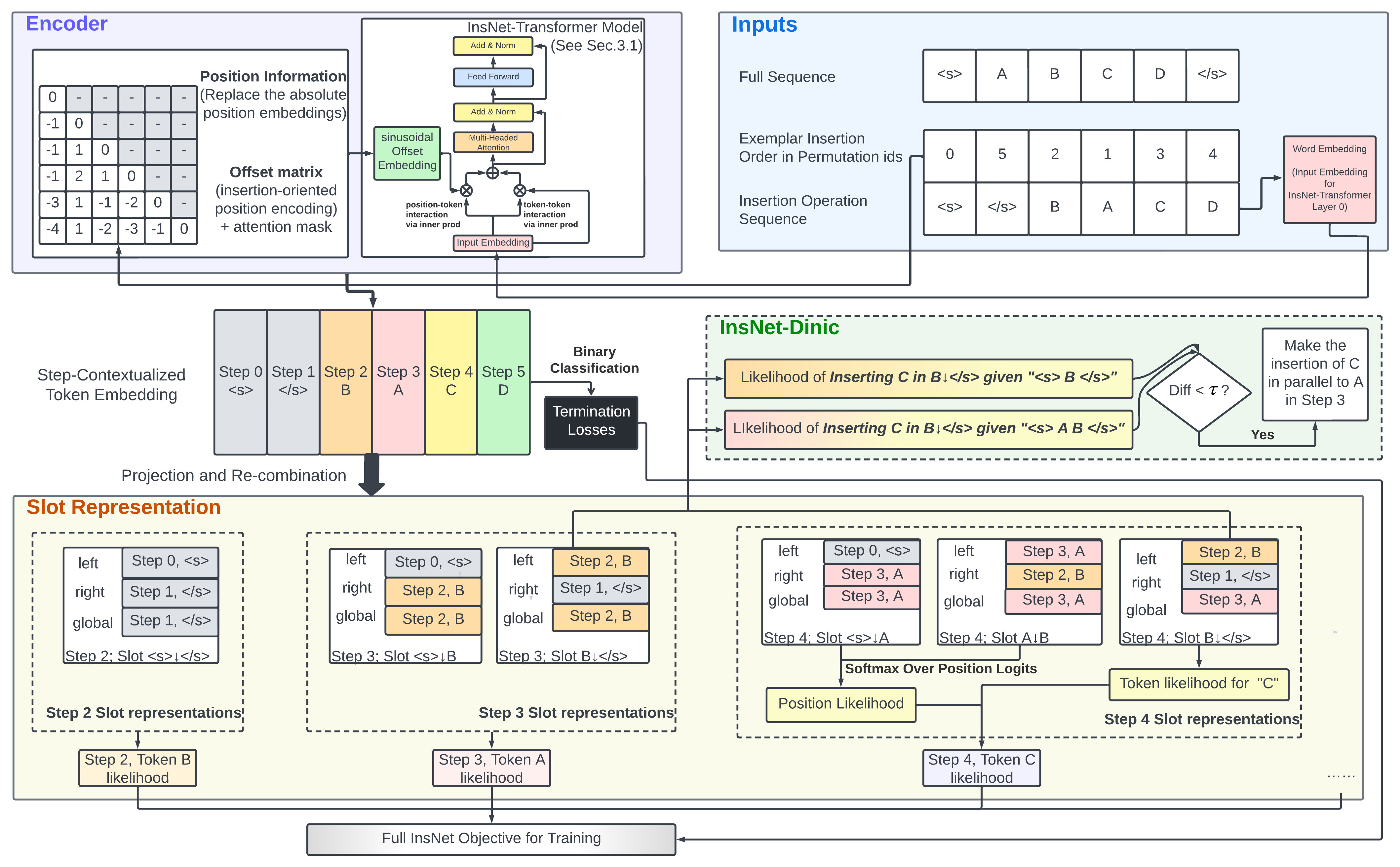}
   \caption{Illustration of the full InsNet model. Given an input sequence with a specified insertion order as illustrated in the [Inputs] panel (the upper right panel in blue), a context encoder (Section~\ref{sec:contextEnc}) illustrated in the [Encoder] panel (the upper left corner in purple) with insertion-oriented position encoding (Section~\ref{sssection:offset}) is applied to the sequence of insertion operations to obtain step-contextualized token embeddings as the transformer outputs. Then, the slot representation module (Section~\ref{ssec:slot}) illustrated in the [Slot Representation] panel (the bottom panel in yellow) compose slot representations for each step from InsNet transformer outputs using its left, right token representation and a global, step-wise token representation. The slot representations are then used to compute the token/position likelihood and also to determine the auto-parallelization in InsNet-Dinic (Section~\ref{ssec:dinic}) illustrated in the [InsNet-Dinic] panel (the middle right panel in green).} 
   \label{fig:full_pipeline}
\vspace{-15pt}
\end{figure}

\section{The \model{} Model}
\label{sec:methodology}
\sidistoryline{Methodological bullet points} 
There are three major components of \model{}: 1) An context encoder based on the transformer architecture~\citep{vaswani2017attention} that uses a novel way to compute \textit{insertion-oriented relative position encoding} to better suit the insertion-based nature of the generation process and enable computation sharing (Section~\ref{sec:contextEnc}); 2) a module to compose expressive \textit{slot representation} for predicting tokens to insert in different slots simultaneously (Section~\ref{ssec:slot}); and 3) an algorithm that adaptively determines the parallelization of the insertion operations to minimize conflicts (Section~\ref{ssec:dinic}).
Figure~\ref{fig:full_pipeline} illustrate the components of the full model.

\subsection{Context Encoder: A Transformer with Insertion-Oriented Relative Position}
\label{sec:contextEnc}
In the introduction, we discussed the challenges in (efficient) computation sharing caused by the volatile position information of the context tokens as shown in Figure~\ref{fig:volatility_pos}. 
We address this problem by empowering the transformer-based context encoder with a distance-aware, insertion-oriented pairwise relative position encoding. The proposed insertion-oriented relative position encoding shares important designs with previous relative position embeddings\citep{dai-etal-2019-transformerxl,yang2019xlnet,shih2019xl}, but differs in how it accurately depicts the spatial layout (i.e., the actual sequential order) of the inserted tokens in an insertion-based generation process.

\sidistoryline{Describe the process with a concrete example to let the reader think what should be the optimal representation (i.e. for each step, we can fully recover the spatial correlation simply by organizing the tokens in the layout defined by such representation). Position volatility already bans the usage of absolute positions in computation sharing, so we no doubt will be using relative positions.} 

\begin{figure}[!b]
  \centering 
  \subfigure[Absolute Positions]{\includegraphics[height=0.22\columnwidth]{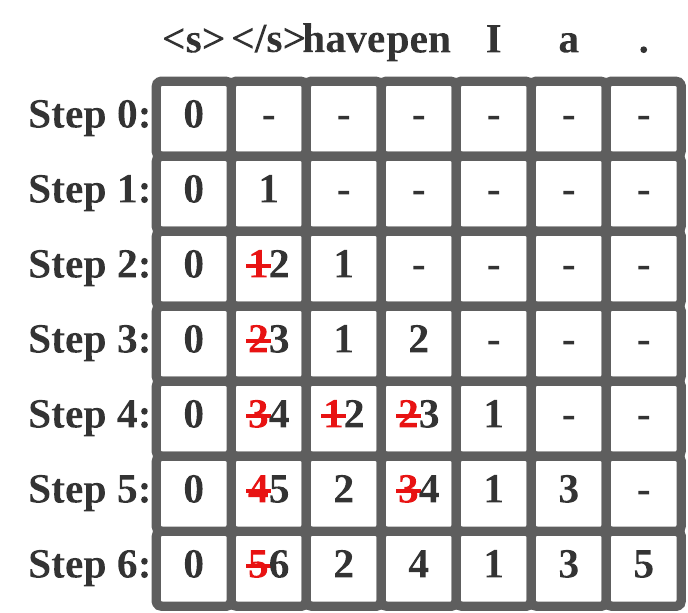}
   \label{fig:offset_matrix_absolute}}  
  \subfigure[Vanilla Relative Positions]{\includegraphics[height=0.22\columnwidth]{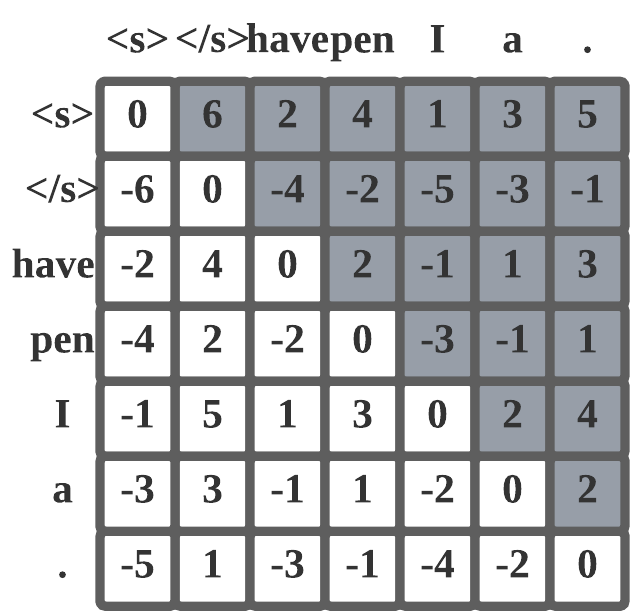}
   \label{fig:offset_matrix_xlnet}} 
  \subfigure[Our Relative Positions]{\includegraphics[height=0.22\columnwidth]{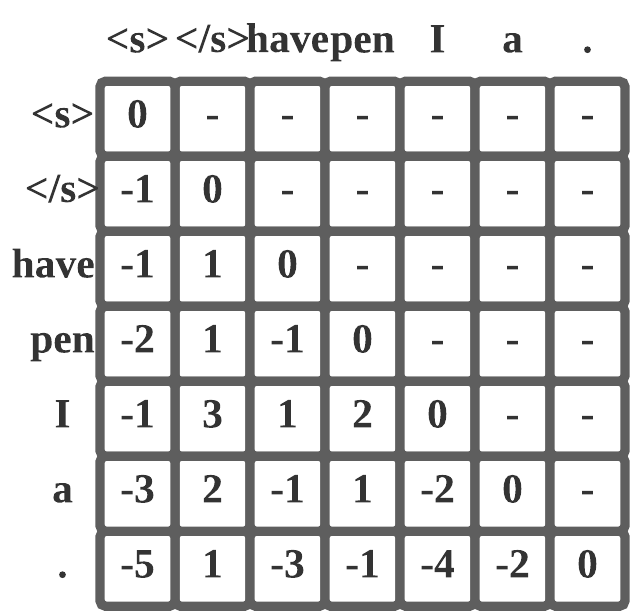}
   \label{fig:offset_matrix_perm}} 
  \subfigure[Ours, Order-Restored]{\includegraphics[height=0.22\columnwidth]{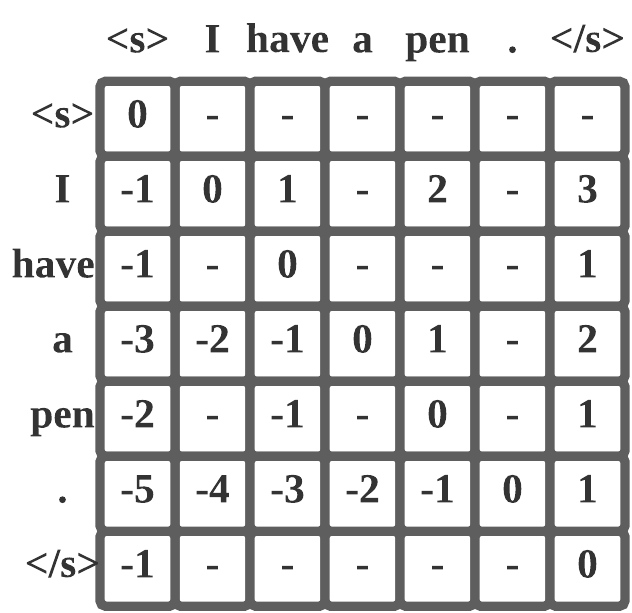}
   \label{fig:offset_matrix_nat}}
  
  \caption{Comparison of different position encodings for an insertion-based generation process. From left to right, we illustrate \ref{fig:offset_matrix_absolute} the volatile absolute positions, \ref{fig:offset_matrix_xlnet} the traditional non-insertion-oriented relative positions, \ref{fig:offset_matrix_perm} the proposed offset matrix presented in the insertion order, and \ref{fig:offset_matrix_nat} the same offset matrix permuted to show how it looks like if we restore the actual (partial-)sequence.} 
  
  \label{fig:offset_matrix}
\end{figure} 

\subsubsection{Offset: Insertion-Oriented Relative Position Encoding}  
\label{sssection:offset}
To illustrate our relative position encoding, considering the partial context (to be completed by further insertions) of \emph{``I have pen''} with an insertion order of \emph{``have pen I''}. To insert an \emph{``a''} in between \emph{``have''} and \emph{``pen''}, the distance vector for token \emph{``a''} against the rest of the context inserted so far should be [(``have'', -1), (``pen'', +1), (``I'', -2)], simplified as [-1, +1, -2].
This relative position encoding clearly defines where the insertion happens by only describing the pairwise spatial relationship between the incoming token and the existing context tokens. 
We can pack the relative position vectors for each insertion step to get a matrix that reflects the relative spatial relation along the trajectory of insertions, with each row corresponding to an insertion step. We name it the \emph{offset} matrix. 

As a concrete example, to generate the sentence \emph{``I have a pen.''} with an insertion order ``\textlangle 
BOS\textrangle ~\textlangle EOS\textrangle ~have pen I a .'', the complete offset matrix is shown in Figure~\ref{fig:offset_matrix_perm}.
The effective elements of this matrix are all in the lower triangular region, making the application of the computation sharing trick for the decoder transformers trivial. 
This reduces the number of context (re-)encoding steps from $n$ in naive insertion-based models~\citep{stern2019insertion} to 1 during training. 

\tinysection{Discussion: Comparisons With Other Positional Encoding Strategies}
To illustrate how the proposed insertion-oriented relative position encoding is different from existing position encoding strategies, 
Figure~\ref{fig:offset_matrix} shows the volatile, un-reusable absolute positions in ~\ref{fig:offset_matrix_absolute}, the traditional non-insertion-oriented relative positions as used in models such as XLNet~\cite{yang2019xlnet} in \ref{fig:offset_matrix_xlnet}, the proposed offset matrix in the insertion order in \ref{fig:offset_matrix_perm}, and the same offset matrix permuted to show the restored actual (partial-)sequence in \ref{fig:offset_matrix_nat}. 
\violet{add discussions about the advantages of ours over other positional encodings.}
We can see that the absolute and the traditional relative positional encoding strategies are not applicable for efficient insertion-based generators. The former requires constant updates of the positional encoding of the existing tokens as the insertion progresses, disabling computation sharing. The latter requires knowledge about the final sequence length at the beginning of the insertion process, which is unrealistic and destroys the flexibility of insertion-based models.
Our relative position encoding, on the other hand, reflects the order of the original sequence. Tokens that are on the left of the current one in the original sequence have a negative relative position to the current one while preserving the insertion order. As is illustrated in Figure~\ref{fig:offset_matrix_nat}, all the ``later inserted tokens'' are masked out and excluded from the previously inserted token's computation.

\begin{figure}[t]
  \centering 
  \includegraphics[width=\columnwidth]{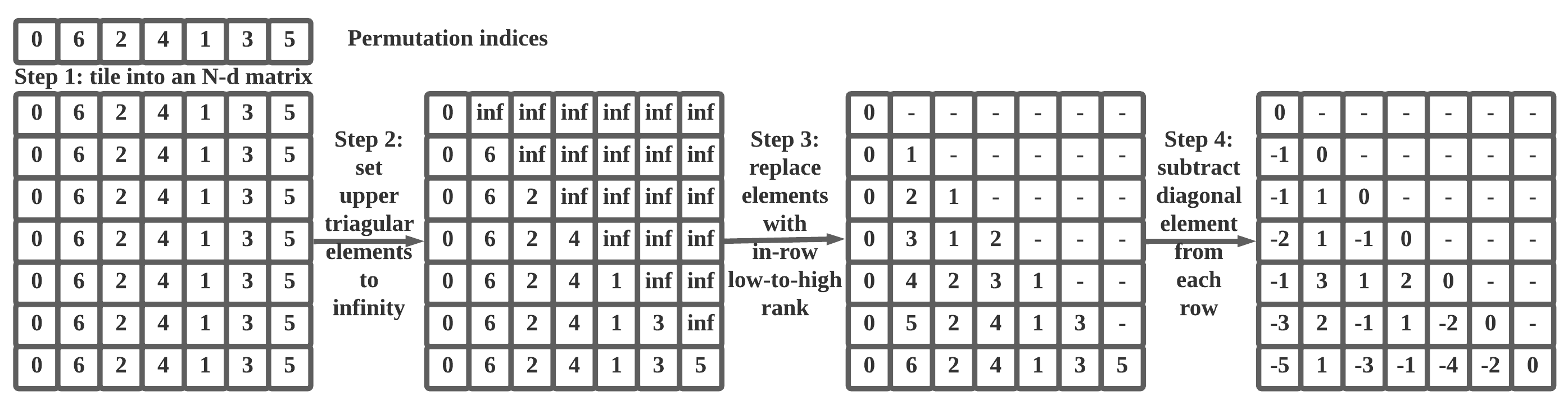}
  \caption{Illustration of the offset compression algorithm. It efficiently transforms the insertion order/permutation indices to an offset matrix. 
  }
  \label{fig:offset_compression}
  \vspace{-20pt}
\end{figure} 

\sidistoryline{now we know what is the desired output of the position encoding mechanism, here we propose an efficient algorithm to produce such encoding in one-pass, giving the full offset matrix} 
\subsubsection{Efficient Computation of Offset Matrix} 
\label{sssection:eff_offset}
During training, given an \emph{insertion order} to generate a sequence, we can pre-compute the offset matrix.
However, naively constructing the offset matrix with element-wise operations takes $O(n^2)$ complexity for each sequence, limiting the training efficiency. 
Therefore we design an efficient algorithm to construct the offset matrix. Assume the \emph{insertion order} is described in the form of absolute position \emph{permutation indices} (i.e., in the previous example, [0, 6, 2, 4, 1, 3, 5]), we design a matrix algorithm called \emph{offset compression} for our purpose.
Figure~\ref{fig:offset_compression} illustrates the algorithm with our exemplar sentence. 
Specifically, we first convert the absolute position vector into a matrix by duplication. 
Then, the upper triangular elements are masked by ``infinity'' to remove their impact in relative position computation because the inserting token should not attend to future to-be-inserted tokens. 
In the third step, each element is replaced by its in-row ranking \emph{i.e.} its absolute position skipping the masked positions. In the final step, each row is baselined by the diagonal element to reflect that the relative position is between the last inserted token (the diagonal element) and previously inserted ones. The algorithm can be efficiently executed as a series of fast matrix operations.



\subsubsection{An \model{} Layer} \sidistoryline{formulate how such offset matrix is incorporated into each layer to make the description self-contained} 
\label{sssection:layer}
We hereby describe how to incorporate the offset matrix into each transformer layer in \model{} for efficient context encoding. In general, we inherit most designs from previous ones with relative position encoding \citep{dai-etal-2019-transformerxl,yang2019xlnet}. 
The encoder panel in Figure~\ref{fig:full_pipeline} (the upper left panel in purple) illustrate the process.
Specifically, for layer number $i=1...N$, given the output embedding $\mat{E}^{i-1}$ from the last layer and the offset matrix $\mat{R}$, the formulation of a transformer layer in an $N$-layer \model{} can be written as:

\vspace{-15pt}
\begin{align*}
    \mat{Q}^i,\mat{K}^i,\mat{V}^i,\mat{P}^i=&\mat{W}_q^i\mat{E}^{i-1},\mat{W}_{k,E}^i\mat{E}^{i-1},\mat{W}_v^i\mat{E}^{i-1}, \mat{W}_{k,R}^i\mat{R}\\
    \mat{A}^i=&\mat{Q}^{i\top}\mat{K}^i + \mat{Q}^{i\top}\mat{P}^i+\mat{u}^{i\top} \mat{K}^i + \mat{v}^{i\top} \mat{P}^i\\
    \mat{V}_{\textnormal{reduced}}^{i}=&\textnormal{Masked-Softmax}(\mat{A}^i)\mat{V}^i\\
    \mat{V}_{\textnormal{skipconn}}^{i}=&\mat{V}_{\textnormal{reduced}}^{i}+\mat{E}^{i-1} \\
    \mat{E}^i=&\textnormal{Feed-Forward}(\mat{V}_{\textnormal{skipconn}}^{i}; \theta^{i}),
\end{align*}
where $\mat{Q}^i,\mat{K}^i,\mat{V}^i,\mat{P}^i$ are query, key, value, position matrices, respectively, for layer $i$. $\mat{u}^i, \mat{v}^i, \mat{W}_q^i, \mat{W}_{k,E}^i, \mat{W}_v^i, \mat{W}_{k,R}^i$ and $\theta^{i}$ are learnable model parameters of each \model{} layer. We denote the input word embeddings as $\mat{E}^{0}$ for notation consistency.
The last layer of transformer in \model{} outputs the step-wise context-aware token embeddings. They will be used to compute slot representations for the insertion steps.

\subsection{Slot Representation and Insertion Prediction} 
\label{ssec:slot}
\sidistoryline{now we have the transformer outputs i.e. the encoding of partial context for every single step in the insertion-based generation trajectory. We need a mechanism to turn it into a representation for each potential target ("slot") of the insertion operation. We make these representation vectors efficiently aggregated from the transformer output. We abandoned the deep aggregation idea, the shallow aggregation seems both efficient and powerful.} 
With the context efficiently encoded, the subsequent step of an insertion-based generator is to predict the next \textit{position and token} to insert. 
Thus, a representation for each potential insertion slot should be computed.  
We hereby show how slot representations can be \emph{aggregated} from \model{} outputs and the potential challenges during this aggregation process.

 

A naive design of the slot representation is to simply concatenate the representation vectors from the left-neighbor and right-neighbor (in the natural observation order) of the slot as is done in prior works~\citep{stern2019insertion}. 
This slot representation can efficiently compute the slot representation for all possible slots in parallel for each time step. 
However, the context encoding we obtained is insertion-order sensitive and unaware of the tokens inserted later in \model{}. 
This naive slot representation thus falls short of capturing the global sequence information. 
As a remedy, we propose to also include the last insertion token's representation to compose the slot representation $\boldsymbol{e}_{i\downarrow j}^{(t)}$. 
Specifically, given the representations of the left-side neighbor $\boldsymbol{e}_{i-}$, the right-side neighbor $\boldsymbol{e}_{+j}$ and the last inserted token $\boldsymbol{e}_{t}$, the slot representation can be computed as:



\vspace{-10pt}
$$\boldsymbol{e}_{i\downarrow j}^{(t)} = \text{LayerNorm}((f_{l}(\boldsymbol{e}_{i-}) \oplus f_{r}(\boldsymbol{e}_{+j})) + \boldsymbol{e}_{t})$$

where $f_{l}$ and $f_{r}$ are linear projections for left-point and right-point representation vectors and $\oplus$ stands for vector concatenation. The slot representation panel in Figure~\ref{fig:full_pipeline} (the bottom panel in yellow) illustrates the process. Note that comparing step 3 and step 4, both compute the slot representation for B $\downarrow$ </s>, but the resulting slot representation changes in response to a new step due to the introduction of the global vector.

The slot representation can then be converted into a probabilistic distribution over the vocabulary to predict the log-likelihood of inserted tokens using a log-linear transformation \emph{i.e.} $\log p(\token_{i\downarrow j}^{(t)}|\boldsymbol{e}_{i\downarrow j}^{(t)})=\text{log-softmax}(\mat{W}_p\boldsymbol{e}_{i\downarrow j}^{(t)} + \boldsymbol{b})_{\token_{i\downarrow j}^{(t)}}$. 
To decide which slot to insert next, a position logit $\alpha_k^{(t)}=w_o^\top \boldsymbol{e}_{i_k\downarrow j_k}^{(t)}$ is computed for each slot $k$ with a linear layer. Then a (log-)soft-max operation is applied on top of the position logits to obtain the log-probability to insert to each candidate slot. 
\emph{i.e.} $\log o(\token_{i_k\downarrow j_k}^{(t)}|\{\boldsymbol{e}^{(t)}_{\downarrow}\})=\text{log-softmax}([\alpha_0^{(t)}, \alpha_1^{(t)}, ...])_k$. 

We apply a binary classification $q(0/1|\cdot)$ on the last inserted token's representation
$\boldsymbol{e}_t$ to decide the termination of the generation at each step. During training, only the final step $q(0/1|\boldsymbol{e}_{n-1})$ is trained to predict 1; all intermediate steps are trained to predict 0. In the following section, we will continue to discuss how to formulate the final objective function of \model{} using these step-wise distributions.

\subsection{Adaptive Parallelization of Insertions} 
\label{ssec:dinic}
\sidistoryline{Stating the legacies from Dinic's Algorithm - layerization of nodes and adjusting the flow based on some score. Here we want this algorithm to wisely parallelize the token so that we cause the least incoherence parallel insertions. We formulate the scores for such incoherence as the discrepancy between the likelihood estimated by the model itself before/after some parallelization proposal. }
In addition to our efforts for better training efficiency, we also propose an algorithm to adaptively parallelize the generation process of \model{} to speed up the decoding while preserving the generation quality. The idea mimics the graph layerization process in the Dinic's algorithm~\citep{dinitz2006dinitz}. 
Specifically, assume that we partition the tokens into different \emph{layers}, such that the tokens within the same layer are \emph{safe} to insert in parallel without affecting the coherence of context. The algorithm aims to determine the partition such that the log-likelihood estimation of the resulting parallel insertion-based model is as close to the sequential insertion-based models as possible. 
The intuition is that the sequential insertion process maximally captures the inter-dependencies between the output tokens. By staying close to the sequential insertion-based model regarding likelihood prediction ability, we speed up the inference without severely sacrificing the likelihood estimation quality. 

\begin{figure}[t]
  \centering 
  
  \subfigure[Initial Layerization]{\includegraphics[height=0.20\columnwidth]{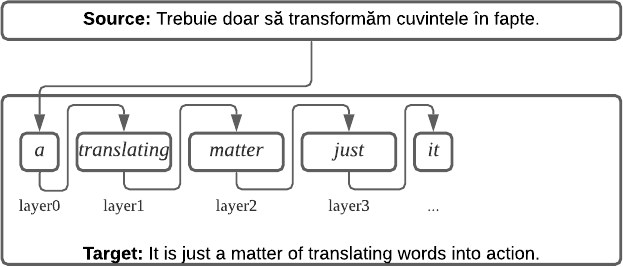}\label{fig:init-layerization}} 
  \subfigure[Merged Layerization]{\includegraphics[height=0.20\columnwidth]{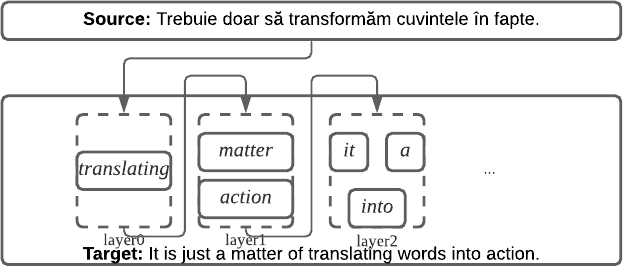}\label{fig:evolve-layerization} }

  \caption{(a) Initialized layerization of tokens after the pre-training stage. It assumes a fully sequential decoding process for now. (b) As the parallelization fine-tuning proceeds, the algorithm gradually merges different layers. Tokens within the same layer are safely parallelizable in decoding without sacrificing log-likelihood over the tolerance threshold $\tau$.}

\end{figure}
 
\sidistoryline{so the model has to be trained as a good likelihood estimator otherwise the its self-prediction of the discrepancy will be less trustworthy}
We initialize the layerization with a sequential insertion order, which is equivalent to single-node layerization, as is illustrated in Figure~\ref{fig:init-layerization}. \sidistoryline{The promotion (or as in Bellman-Ford algorithm, the "relaxation") operation that only parallelize tokens that cause neglectable (as indicated by $\tau$) likelihood loss}
Starting from this initialization, we perform \model-Dinic to gradually evolve this fully-sequential layerization into a non-autoregressive, parallelized layerization, as shown in Figure~\ref{fig:evolve-layerization}. We \emph{promote} a token $\token_\leftarrow$ from its respective slot $(i^{l}, j^{l})$ in layer $l$ to $(i^{l-1}, j^{l-1})$ in the previous layer $l-1$ if two conditions are satisfied: 1) In layer $l-1$, slot $(i^{l-1}, j^{l-1})$ does not yet have an assigned insertion; 2) $\log p(\token_\leftarrow|\boldsymbol{e}_{i^l\downarrow j^l}^{(l)}) - \log p(\token_\leftarrow|\boldsymbol{e}_{i^{l-1}\downarrow j^{l-1}}^{(l-1)}) \leq \tau$.

Here $\tau$ is a hyper-parameter that controls how much to parallel. 
Intuitively, larger $\tau$ usually indicates more tolerance for token likelihood loss caused by parallel decoding, \emph{i.e.} more parallelization with tradeoff on likelihood estimation accuracy. 
The choice of $\tau$ is highly task/data-dependent as it is a value associated with the likelihood. 
Note that the \emph{promotion} of tokens only affects the slot representation computation and the insertion likelihood calculation of the respective insertion. 
When encoding the context (as is discussed in Section~\ref{sec:contextEnc}), we always treat it as if these insertion operations are performed sequentially.

\sidistoryline{the resulting formulation of the objective for \model{}/\model-Dinic} 
For a target sequence $\sequence_{<n}=[\token_0, \token_1, ..., \token_{n-1}]$, after the layerization algorithm, if token $\token_{i\downarrow j}$ is layerized in the $l$-th layer and to be inserted in the slot $(i, j)$, we denote it as $\token^l_{i\downarrow j}$. 
Suppose we have $m$ effective layers in total. Denote the corresponding representation for slot $(i,j)$ given available context in layer $l$ as $\boldsymbol{e}^{(l)}_{i\downarrow j}$ and the transformer's encoding of the partial context until the end of layer $l$ as $\boldsymbol{e}^{(l)}_{-1}$.
We denote the token likelihood, position likelihood and termination likelihood of the token $\token^l_{i\downarrow j}$ to be $p(\token^l_{i\downarrow j} | \boldsymbol{e}^{(l)}_{i\downarrow j})$, $o(\token^l_{i\downarrow j} |\{ \boldsymbol{e}^{(l)}_{\downarrow}\})$ and $q(0/1| \boldsymbol{e}^{(l)}_{-1})$, respectively. The general objective of \model{} parameterized by $\phi$ can be formulated as:

\vspace{-10pt}
\begin{equation}
    \mathcal{L}(\sequence, \phi) = -\mathop{\sum}_{l=0}^{m-1} \mathop{\sum}_{\token_{i\downarrow j}^l}[ \log p_\phi(\token_{i\downarrow j}^l | \boldsymbol{e}^{(l)}_{i\downarrow j}) + \log o_\phi(\token_{i\downarrow j}^l | \{\boldsymbol{e}^{(l)}_{\downarrow}\})] \nonumber -\mathop{\sum}_{l =0}^{m} \log q_\phi(\mathbf{1}(l=m)|\boldsymbol{e}^{(l)}_{-1})
\end{equation}


\begin{table*}[t]
  \centering
  \caption{Performance comparison on Yelp Review and News datasets. For Levenshtein Transformer, insertion and deletion stages are both counted in \# of decoding-time iterations. It's non-trivial to do unsupervised lexically constrained text generation with auto-regressive models. To work around this, we implemented a Plan-And-Write\citep{yao2019plan} style auto-regressive transformer-based model for better reference. Models with a star mark * are re-implemented by us. Other baseline results are directly taken from the original papers.}
  \small
  \resizebox{\textwidth}{!}{
  \begin{tabular}{lcccccc}
    \toprule
    \dual{Model} & \multicolumn{3}{c}{Yelp Review} & \multicolumn{3}{c}{News}\\
    & BLEU-2/4 & NIST-2/4 & \# Dec. Steps & BLEU-2/4 & NIST-2/4 & \# Dec. Steps \\
   \midrule Auto-regressive Transformer & \dual{16.68/5.46} & \dual{2.79/2.86} & \dual{39.24} & \dual{8.79/2.40} & \dual{1.65/1.67} & \dual{36.74}  \\
   (Plan-And-Write-static, \citet{yao2019plan})\\
   \midrule
   NMSTG \citep{welleck2019non} & 10.06/1.92 & 1.11/1.12 & 27.92 & 10.67/1.58 & 2.70/2.70 & 27.85 \\
   InDIGO* \citep{gu2019insertion} & \dual{16.14/4.63} & \dual{3.08/3.10} & \dual{45.63}  & \dual{13.89/3.62} & \dual{3.08/3.10} & \dual{26.78} \\
   (w/ Searched Adaptive Order) \\
   \midrule
   Levenshtein Transformer &  \dual{14.84/3.96} & \dual{2.84/2.89} & \dual{14.28} & \dual{11.76/1.89} &  \dual{2.64/2.71} & \dual{16.13} \\ 
   (Parallel Decoding,  \citep{gu2019levenshtein}) \\ 
   \midrule
   InsT-POINTER-Base (BERT init)  & 11.48/2.16 & 2.15/2.15 & 6.00 & 12.13/1.63 & 2.90/2.80 & 6.00  \\
  InsT-POINTER-Base (BERT init+Wiki)  & 15.63/3.32 & 3.27/3.30 & 6.00 & 13.01/2.51 & 3.04/3.06 & 6.00  \\
   InsT-POINTER-Large (BERT init+Wiki)  & 16.78/3.79 & 3.49/3.53 & 6.00 & 14.04/3.04 & \textbf{3.28/3.30} & 6.00  \\
   \midrule
   \model{}  (Ours, Fully-Sequential) & \textbf{19.36/5.78} & \textbf{3.51/3.54} & 48.73 & \textbf{16.31/4.96} & 3.10/3.13 & 32.69 \\ 
   \model-uniform (Ours) & 12.31/2.30 & 2.19/2.17 & 7.00 & 12.89/2.01 & 2.99/2.90 & 7.00 \\
   \model-Dinic  (Ours, $\tau=10.0$) & 16.73/4.35 & 3.19/3.20 & 11.83 & 14.13/3.75 & 2.97/3.00 & 8.13 \\
   
  \bottomrule
  \end{tabular}
  }
  
  \label{tab:lex}
\end{table*}

\section{Experiments}
\sidistoryline{Simplified experiment setup: spectral study of the quality-efficiency trade-off on Yelp and News; NMT} 
We demonstrate the efficiency, flexibility, and model capability of \model{} with two sets of experiments. 1) To show \model's performance and the flexibility to switch between sequential and parallel decoding on datasets with high inter-token dependency (i.e., less suitable for parallel decoding). We follow the setup in \citet{zhang2020pointer} and address the unsupervised lexically constrained text generation problem on two datasets Yelp Review and News. 2) To further verify the effectiveness of \model-Dinic, we evaluate \model{} as a (partially) non-autoregressive machine translation model with parallel decoding on three classical datasets: WMT Ro-En, WMT En-De and WAT En-Ja.
The general setup about how hyper-parameters are determined can be found in appendix\ref{sec:appendix}.

\subsection{Lexically Constrained Generation}  
\sidistoryline{Boring details about data stats} \tinysection{Experimental Setup} Yelp Review dataset consists of 160K training sequences, 10K sequences for validation and 1k test sequences. News dataset consists of 268586 sentences in total, of which 10k are randomly selected as validation set, 1k for testing. YAKE \footnote{https://github.com/LIAAD/yake} is performed to the test split of each dataset to extract lexical constraints.

\begin{figure}[h]
\vspace{-15pt}
  \centering
  \subfigure[Yelp Review]{\includegraphics[width=0.45\columnwidth]{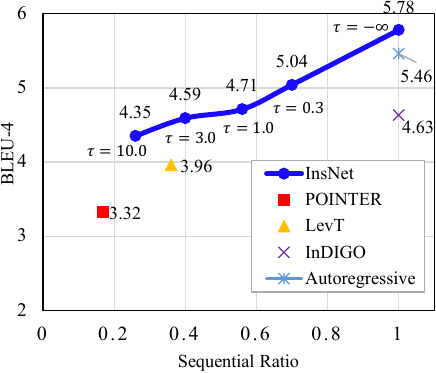}\label{fig:yelp_spec} }
  \subfigure[News]{\includegraphics[width=0.45\columnwidth]{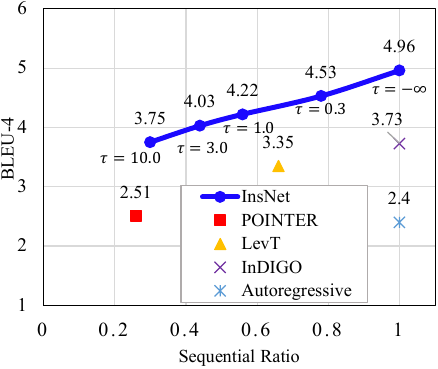}\label{fig:news_spec} }
  
  \caption{Illustration of the spectral study results in terms of BLEU-4/Sequential Ratio in decoding on Yelp Review and News dataset. Sequential Ratio is computed by $\frac{Avg. \text{\# Dec.Steps}}{Avg. \text{Length}}$}
\vspace{-15pt}
  
\end{figure}

\sidistoryline{how we select $\tau$} We vary the hyperparameter $\tau$ in the range of $\{10.0, 3.0, 1.0, 0.3, -\infty(\text{fully-sequential})\}$ and collect the results on both datasets. For position prediction, we are inserting into slots with positions lying in the top $70\%$ of the position distribution mass. For token prediction, we are doing top-$\{1, 1, 3, 3, 5\}$ sampling over the vocabulary distribution. The results are shown in Table~\ref{tab:lex}, Figure~\ref{fig:yelp_spec} and \ref{fig:news_spec}

\tinysection{Discussion} 1) \model{} is able to significantly outperform most previous works on the quality-latency trade-off spectrum. With more sequential flavor (\emph{i.e.} smaller $\tau$), it is generally to achieve the new state-of-the-art in generation quality. Compared to a Plan-And-Write style autoregressive baseline, \model{} guarantees the perfect incorporation of keywords due to its insertion-based nature. Also, its improved performance is consistent on both datasets. 2) When setting $\tau$ larger to encourage more parallelization of decoding, \model{} is mostly able to parallelize the insertions with good preservation of the generative quality. Compared to the vanilla uniformly-parallel setup (\model{}-uniform), \model-Dinic significantly improved the effectiveness of parallel decoding.
 
\tinysection{Generation sample} We also show two samples of the generation process in Table~\ref{tab:body_generative_traj}. It is interesting to observe that the model tends to insert punctuation and function words (e.g., Articles and Prepositions) first, and then more concrete content words.

\begin{table}[]
    \centering
    \scriptsize
    \renewcommand{\tabcolsep}{0.5mm}
    \caption{Generation examples with insertion steps. Every row shows 3 consecutive steps with \colorbox{orange!30}{orange}, \colorbox{green!30}{green}, \colorbox{blue!30}{blue} represents the first, second, and third insertion, respectively.}
    \begin{tabular}{l|p{6cm}|p{6.3cm}}
    \toprule
    \textbf{Input}  & \textbf{day decided started focus on} & \textbf{local group hurt rule out} \\ \hline
    \textbf{Step 3} & \colorbox{green!30}{the} day decided \colorbox{blue!30}{to} started focus on \colorbox{orange!30}{.} & \colorbox{blue!30}{the} local group hurt rule out \colorbox{orange!30}{of} \colorbox{green!30}{.} \\ \hline
    \textbf{Step 6} & the day \colorbox{orange!30}{,} \colorbox{green!30}{he} decided to \colorbox{blue!30}{get} started focus on . &  the local group hurt \colorbox{blue!30}{the} rule out of \colorbox{orange!30}{the} \colorbox{green!30}{of} .\\ \hline
    \textbf{Step 9} & \colorbox{orange!30}{on} the day , he decided to get started focus on \colorbox{green!30}{the} \colorbox{blue!30}{court} . &  the local group hurt the \colorbox{blue!30}{government} rule out of the of \colorbox{orange!30}{the} \colorbox{green!30}{year} .\\ \hline
    \textbf{Step 12} & \colorbox{orange!30}{but} on the \colorbox{blue!30}{next} day , he decided to get started \colorbox{green!30}{to} focus on the court . &  the local group \colorbox{orange!30}{has} hurt the government \colorbox{blue!30}{to} rule out of the of the \colorbox{green!30}{last} year .\\ \hline
    \textbf{Step 15} & but \colorbox{orange!30}{,} on \colorbox{blue!30}{the} next day , he decided to get started to focus on the court \colorbox{green!30}{for} the . &  the local group has \colorbox{green!30}{been} hurt the government to rule out of \colorbox{blue!30}{for} the \colorbox{orange!30}{rest} of the last year . \\ \hline
    \textbf{Step 17} & but , on the next day , he decided to get started to focus on the court for the \colorbox{green!30}{first} \colorbox{orange!30}{time} . &  the local group has been hurt \colorbox{orange!30}{by} the government to rule out of \colorbox{green!30}{support} for the rest of the last year .\\ \bottomrule
    \end{tabular}
    \label{tab:body_generative_traj}
\end{table}

\subsection{Non-autoregressive Machine Translation} 
\sidistoryline{How we trained the model on NMT} We train \model-Dinic models with $\tau=2.00$ for extended investigation of the performance on machine learning problems. The results in Table~\ref{tab:machine_translation} show that, \model-Dinic is able to achieve comparable or even better performance compared to previous state-of-the-art non-autoregressive (insertion-based) machine translation models in terms of generation quality and latency.

\begin{table}[b]

  \centering
  \small
  \renewcommand{\tabcolsep}{1mm}
  \caption{Machine translation evaluation. Each generation iteration of Levenshtein Transformer requires at least two full executions of the transformer model. Results with a star mark * are collected from our re-implementation. Other baseline results are directly taken from the original papers. The results for a vanilla transformer is taken from the LevT paper~\citep{gu2019levenshtein}. }
\begin{tabular}{lccccc}
    \toprule  
    Model &  Ro-En &  En-De & En-Ja & \#Dec Step. & Latency\\
   \midrule
   InDIGO-SAO (w/o KD) \citep{gu2019insertion} & 32.47 & 26.14* & 40.87* & $n$ & 516ms\\
   InDIGO-random (w/o KD) \citep{gu2019insertion} &  20.20 & 17.48* & 23.91* & $n$ & 502ms\\
   \midrule
   InsT-uniform (+KD) \citep{stern2019insertion} & 28.52 & 26.72 & 41.89* & $\Theta(\log n) \leq 10$ & 107ms \\
   InsT-binary ($\tau=0.5$, +KD) \citep{stern2019insertion} & 30.66 & 27.41 & 42.17* & $\Theta(\log n) \leq 10$ & 92ms \\
   LevT (+KD)\citep{gu2019levenshtein} & 33.26 & 27.27 & 42.36 & $\Theta(\log n) \leq 2 \times 10$ & 116ms\\  %
   \midrule
   Vanilla Transformer (w/o KD)   & 32.30 & 27.17 & 43.68 & $n$ & 389ms\\
   \midrule
   \model-uniform & 29.13 & 26.45 & 41.67 & $\Theta(\log n) \leq 10$ & 92ms \\
   \model-Dinic  (Ours, $\tau=2.00$) & 33.41 & 27.36 & 43.71 & $\Theta(\log n) \approx 15.8$ & 105ms\\
   + KD & \textbf{33.91} & \textbf{28.05} & \textbf{44.10} & $\Theta(\log n) \approx 16.1$ & 103ms\\
   
  \bottomrule
  \end{tabular}
  \label{tab:machine_translation}
\end{table}

\begin{table}[h]
  \centering
  \small
\caption{Empirical \& theoretical comparative study of different algorithms' training efficiency. \# of Steps/Seq means during training how many time the transformer needs to be executed per sequence.
}
  \begin{tabular}{lcc}
    \toprule
    Model & $T_{train}$/Epoch & \# of Steps/Seq  \\ 
   \midrule 
   Auto-regressive Decoder Transformer & 35min48s & 1 \\
   InDIGO-SAO & 58min36s + 2h12min(SAO) & 1 + $O(n)$\\
   \midrule
   Insertion Transformer-Sequential & 26h31min & $O(n)$ \\
   Levenshtein Transformer-Sequential & 37h24min & $O(n)$ \\
   \midrule
   Insertion Transformer & 8h24min & $\Theta(\log n)$ \\
   Levenshtein Transformer & 16h33min & $\Theta(\log n)$ \\
   InsT-POINTER & 9h36min & $\Theta(\log n)$\\
   \midrule
   \model{}/\model-Dinic & 1h12min & 1 \\
   
  \bottomrule
  \end{tabular}
  \label{tab:time_ana}
\end{table}
\subsection{Training Time Analysis} 

\sidistoryline{empirical training efficiency analysis. give the reader a more direct and intuitive understanding about how fast O(1) training can be, compared to $\Theta(n)$; and why $O(n)$ is intolerable} To show the training efficiency of the proposed model, we hereby do an empirical and theoretical analysis on the candidate models' training procedure. See Table~\ref{tab:time_ana}. For empirical results, the statistics are collected on the Yelp Review unsupervised lexically constrained text generation problem. All results are collected on a single NVIDIA RTX3090 GPU. The transformer Seq2seq baseline is trained as a Plan-And-Write model. POINTER's training time per epoch is calculated with the official implementation with mixed precision supported by the NVIDIA APEX library \footnote{https://github.com/dreasysnail/POINTER}.

\tinysection{Discussion} The results in Table~\ref{tab:time_ana} confirm the significant improvement in training efficiency of \model{} over most previous baselines. In addition, this concretely shows that it is not practically affordable to obtain a sequential insertion-based generator directly with InsT/LevT. Note that SAO of InDIGO is an offline, inference-only algorithm that does not require any computation of gradients. Thus its constant factor is significantly smaller than other $O(n)$ gradient-propagating procedures. According to its original paper and our previous experiments, InDIGO practically needs this process to achieve reasonable performance.

\subsection{Ablation Study on the Effectiveness of Components}
Since the model components of InsNet are closely entangled, it is nontrivial to do ablation study. We design experiments to show that 1). The proposed offset matrix is a powerful insertion-oriented position encoding with significantly better capacity. We show that this particularly helps the model learn to terminate the sequence in correct timings. To study this, we truncate the offset matrix to range $[-1.0, 1.0]$ (InsNet-truncated) so that it is similar to InDIGO's position encoding, and 2) the global representation is necessary for computing the slot representation in \model. It in general helps to improve the sequence termination performance, and leads to convergence to better local optimum. We report the results of an ablated version of InsNet without global representation (InsNet-noglobal).
    
The termination NLL (see Sec 3.3 for its definition) and BLEU scores for each model can be found in appendix Table~\ref{appentab:ablation_quan}. We also show some random samples from the ablated variants and the failure cases can be find in appendix Table~\ref{appentab:ablation}.
\newcommand{\rb}[1]{\textcolor{MyRed}{\textbf{#1}}}
\newcommand{\bb}[1]{\textcolor{MyDarkBlue}{#1}}
\newcommand{\pb}[1]{\textcolor{MyPurple}{\textbf{#1}}}
        
\section{Conclusion \& Future Work}

\sidistoryline{conclusion} We propose \model{}, an efficient and performant insertion-based generator that supports sequential and parallel decoding. 
Experiments on two unsupervised lexically constrained text generation datasets and three machine translation datasets show the advantages of \model{} over previous methods.

Future work can explore obtaining a large-scale pre-trained version of \model{} for further fine-tuning under different downstream scenarios. We anticipate such insertion-based models to have better compositional generalizability and controllability. 

\section*{Acknowledgement}
We thank I-Hung Hsu, Dr. Rujun "RJ" Han, Te-Lin Wu, Kai-Wei Chang, Sarik Ghazarian, Alexander Spangher, Yining Hong, Mingyu Derek Ma and all other members from PlusLabNLP/UCLANLP group for their participation in initial discussions and comments on paper writing. We would like to thank Huggingface for their great work of the Transformers project. 
The work is partially supported by a Meta SRA and an Amazon research gift.


\bibliography{anthology,insnet}

\begin{thebibliography}{33}
\expandafter\ifx\csname natexlab\endcsname\relax\def\natexlab#1{#1}\fi

\bibitem[{Bahdanau et~al.(2015)Bahdanau, Cho, and Bengio}]{dzmitry2015neural}
Dzmitry Bahdanau, {Kyung Hyun} Cho, and Yoshua Bengio. 2015.
\newblock Neural machine translation by jointly learning to align and
  translate.
\newblock 3rd International Conference on Learning Representations, ICLR 2015 ;
  Conference date: 07-05-2015 Through 09-05-2015.

\bibitem[{Brown et~al.(2020)Brown, Mann, Ryder, Subbiah, Kaplan, Dhariwal,
  Neelakantan, Shyam, Sastry, Askell et~al.}]{brown2020gpt3}
Tom~B Brown, Benjamin Mann, Nick Ryder, Melanie Subbiah, Jared Kaplan, Prafulla
  Dhariwal, Arvind Neelakantan, Pranav Shyam, Girish Sastry, Amanda Askell,
  et~al. 2020.
\newblock Language models are few-shot learners.
\newblock \emph{arXiv preprint arXiv:2005.14165}.

\bibitem[{Dai et~al.(2019)Dai, Yang, Yang, Carbonell, Le, and
  Salakhutdinov}]{dai-etal-2019-transformerxl}
Zihang Dai, Zhilin Yang, Yiming Yang, Jaime Carbonell, Quoc Le, and Ruslan
  Salakhutdinov. 2019.
\newblock \href {https://doi.org/10.18653/v1/P19-1285} {Transformer-{XL}:
  Attentive language models beyond a fixed-length context}.
\newblock In \emph{Proceedings of the 57th Annual Meeting of the Association
  for Computational Linguistics}, pages 2978--2988, Florence, Italy.
  Association for Computational Linguistics.

\bibitem[{Dinitz(2006)}]{dinitz2006dinitz}
Yefim Dinitz. 2006.
\newblock Dinitz’algorithm: The original version and even’s version.
\newblock In \emph{Theoretical computer science}, pages 218--240. Springer.

\bibitem[{Goldfarb-Tarrant et~al.(2020)Goldfarb-Tarrant, Chakrabarty,
  Weischedel, and Peng}]{goldfarb2020content}
Seraphina Goldfarb-Tarrant, Tuhin Chakrabarty, Ralph Weischedel, and Nanyun
  Peng. 2020.
\newblock Content planning for neural story generation with aristotelian
  rescoring.
\newblock In \emph{the 2020 Conference on Empirical Methods in Natural Language
  Processing (EMNLP)}, pages 4319--4338.

\bibitem[{Gu et~al.(2019{\natexlab{a}})Gu, Liu, and Cho}]{gu2019insertion}
Jiatao Gu, Qi~Liu, and Kyunghyun Cho. 2019{\natexlab{a}}.
\newblock Insertion-based decoding with automatically inferred generation
  order.
\newblock \emph{Transactions of the Association for Computational Linguistics},
  7:661--676.

\bibitem[{Gu et~al.(2019{\natexlab{b}})Gu, Wang, and Zhao}]{gu2019levenshtein}
Jiatao Gu, Changhan Wang, and Jake Zhao. 2019{\natexlab{b}}.
\newblock Levenshtein transformer.
\newblock \emph{arXiv preprint arXiv:1905.11006}.

\bibitem[{Han et~al.(2022)Han, Chen, Tian, and Peng}]{han2022go}
Rujun Han, Hong Chen, Yufei Tian, and Nanyun Peng. 2022.
\newblock Go back in time: Generating flashbacks in stories with event temporal
  prompts.
\newblock In \emph{2022 Annual Conference of the North American Chapter of the
  Association for Computational Linguistics (NAACL)}.

\bibitem[{He et~al.(2019)He, Peng, and Liang}]{he2019pun}
He~He, Nanyun Peng, and Percy Liang. 2019.
\newblock Pun generation with surprise.
\newblock In \emph{2019 Annual Conference of the North American Chapter of the
  Association for Computational Linguistics (NAACL-HLT 2019)}, volume~1.

\bibitem[{Hempelmann(2008)}]{hempelmann2008humor}
Christian~F Hempelmann. 2008.
\newblock Computational humor: Beyond the pun?
\newblock \emph{The Primer of Humor Research. Humor Research}, 8:333--360.

\bibitem[{Li et~al.(2017{\natexlab{a}})Li, Monroe, Shi, Jean, Ritter, and
  Jurafsky}]{li2017adversarial}
Jiwei Li, Will Monroe, Tianlin Shi, S{\'e}bastien Jean, Alan Ritter, and Dan
  Jurafsky. 2017{\natexlab{a}}.
\newblock Adversarial learning for neural dialogue generation.
\newblock \emph{arXiv preprint arXiv:1701.06547}.

\bibitem[{Li et~al.(2017{\natexlab{b}})Li, Chen, Li, Gao, and
  Celikyilmaz}]{li2017end}
Xiujun Li, Yun-Nung Chen, Lihong Li, Jianfeng Gao, and Asli Celikyilmaz.
  2017{\natexlab{b}}.
\newblock End-to-end task-completion neural dialogue systems.
\newblock \emph{arXiv preprint arXiv:1703.01008}.

\bibitem[{Liu et~al.(2020)Liu, Gu, Goyal, Li, Edunov, Ghazvininejad, Lewis, and
  Zettlemoyer}]{liu2020multilingual}
Yinhan Liu, Jiatao Gu, Naman Goyal, Xian Li, Sergey Edunov, Marjan
  Ghazvininejad, Mike Lewis, and Luke Zettlemoyer. 2020.
\newblock Multilingual denoising pre-training for neural machine translation.
\newblock \emph{Transactions of the Association for Computational Linguistics},
  8:726--742.

\bibitem[{Manurung et~al.(2000)Manurung, Ritchie, and
  Thompson}]{manurung2000poetry}
Hisar Manurung, Graeme Ritchie, and Henry Thompson. 2000.
\newblock Towards a computational model of poetry generation.
\newblock Technical report, The University of Edinburgh.

\bibitem[{Mittal et~al.(2022)Mittal, Tian, and Peng}]{Mittal2022ambipun}
Anirudh Mittal, Yufei Tian, and Nanyun Peng. 2022.
\newblock Ambipun: Generating humorous puns with ambiguous context.
\newblock In \emph{2022 Annual Conference of the North American Chapter of the
  Association for Computational Linguistics (NAACL), short}.

\bibitem[{Paszke et~al.(2019)Paszke, Gross, Massa, Lerer, Bradbury, Chanan,
  Killeen, Lin, Gimelshein, Antiga et~al.}]{paszke2019pytorch}
Adam Paszke, Sam Gross, Francisco Massa, Adam Lerer, James Bradbury, Gregory
  Chanan, Trevor Killeen, Zeming Lin, Natalia Gimelshein, Luca Antiga, et~al.
  2019.
\newblock Pytorch: An imperative style, high-performance deep learning library.
\newblock \emph{Advances in neural information processing systems},
  32:8026--8037.

\bibitem[{Radford et~al.(2018)Radford, Narasimhan, Salimans, and
  Sutskever}]{radford2018gpt1}
Alec Radford, Karthik Narasimhan, Tim Salimans, and Ilya Sutskever. 2018.
\newblock Improving language understanding by generative pre-training.

\bibitem[{Radford et~al.(2019)Radford, Wu, Child, Luan, Amodei, and
  Sutskever}]{radford2019gpt2}
Alec Radford, Jeffrey Wu, Rewon Child, David Luan, Dario Amodei, and Ilya
  Sutskever. 2019.
\newblock Language models are unsupervised multitask learners.
\newblock \emph{OpenAI blog}, 1(8):9.

\bibitem[{Shaw et~al.(2018)Shaw, Uszkoreit, and Vaswani}]{shaw2018self}
Peter Shaw, Jakob Uszkoreit, and Ashish Vaswani. 2018.
\newblock Self-attention with relative position representations.
\newblock \emph{arXiv preprint arXiv:1803.02155}.

\bibitem[{Shih et~al.(2019)Shih, Chang, and Yang}]{shih2019xl}
Yong-Siang Shih, Wei-Cheng Chang, and Yiming Yang. 2019.
\newblock Xl-editor: Post-editing sentences with xlnet.
\newblock \emph{arXiv preprint arXiv:1910.10479}.

\bibitem[{Stern et~al.(2019)Stern, Chan, Kiros, and
  Uszkoreit}]{stern2019insertion}
Mitchell Stern, William Chan, Jamie Kiros, and Jakob Uszkoreit. 2019.
\newblock Insertion transformer: Flexible sequence generation via insertion
  operations.
\newblock In \emph{International Conference on Machine Learning}, pages
  5976--5985. PMLR.

\bibitem[{Sun et~al.(2022)Sun, Narayan-Chen, Oraby, Gao, Chung, Huang, Liu, and
  Peng}]{sun2022context}
Jiao Sun, Anjali Narayan-Chen, Shereen Oraby, Shuyang Gao, Tagyoung Chung, Jing
  Huang, Yang Liu, and Nanyun Peng. 2022.
\newblock Context-situated pun generation.
\newblock In \emph{Proceedings of the 2022 Conference on Empirical Methods in
  Natural Language Processing (EMNLP)}.

\bibitem[{Tan et~al.(2020)Tan, Yang, AI-Shedivat, Xing, and
  Hu}]{tan2020progressive}
Bowen Tan, Zichao Yang, Maruan AI-Shedivat, Eric~P Xing, and Zhiting Hu. 2020.
\newblock Progressive generation of long text.
\newblock \emph{arXiv preprint arXiv:2006.15720}.

\bibitem[{Tian et~al.(2022)Tian, Arun~Sheth, and Peng}]{tian2022unified}
Yufei Tian, Divyanshu Arun~Sheth, and Nanyun Peng. 2022.
\newblock A unified framework for pun generation with humor principles.
\newblock In \emph{Proceedings of the 2022 Conference on Empirical Methods in
  Natural Language Processing (EMNLP)}.

\bibitem[{Tian and Peng(2022)}]{tian2022sonnet}
Yufei Tian and Nanyun Peng. 2022.
\newblock Zero-shot sonnet generation with discourse-level planning and
  aesthetics features.
\newblock In \emph{2022 Annual Conference of the North American Chapter of the
  Association for Computational Linguistics (NAACL)}.

\bibitem[{Vaswani et~al.(2017)Vaswani, Shazeer, Parmar, Uszkoreit, Jones,
  Gomez, Kaiser, and Polosukhin}]{vaswani2017attention}
Ashish Vaswani, Noam Shazeer, Niki Parmar, Jakob Uszkoreit, Llion Jones,
  Aidan~N Gomez, Lukasz Kaiser, and Illia Polosukhin. 2017.
\newblock Attention is all you need.
\newblock In \emph{NIPS}.

\bibitem[{Vinyals et~al.(2015)Vinyals, Toshev, Bengio, and
  Erhan}]{vinyals2015show}
Oriol Vinyals, Alexander Toshev, Samy Bengio, and Dumitru Erhan. 2015.
\newblock Show and tell: A neural image caption generator.
\newblock In \emph{Proceedings of the IEEE conference on computer vision and
  pattern recognition}, pages 3156--3164.

\bibitem[{Welleck et~al.(2019)Welleck, Brantley, Iii, and Cho}]{welleck2019non}
Sean Welleck, Kiant{\'e} Brantley, Hal~Daum{\'e} Iii, and Kyunghyun Cho. 2019.
\newblock Non-monotonic sequential text generation.
\newblock In \emph{International Conference on Machine Learning}, pages
  6716--6726. PMLR.

\bibitem[{Wolf et~al.(2020)Wolf, Debut, Sanh, Chaumond, Delangue, Moi, Cistac,
  Rault, Louf, Funtowicz, Davison, Shleifer, von Platen, Ma, Jernite, Plu, Xu,
  Scao, Gugger, Drame, Lhoest, and Rush}]{wolf-etal-2020-transformers}
Thomas Wolf, Lysandre Debut, Victor Sanh, Julien Chaumond, Clement Delangue,
  Anthony Moi, Pierric Cistac, Tim Rault, Rémi Louf, Morgan Funtowicz, Joe
  Davison, Sam Shleifer, Patrick von Platen, Clara Ma, Yacine Jernite, Julien
  Plu, Canwen Xu, Teven~Le Scao, Sylvain Gugger, Mariama Drame, Quentin Lhoest,
  and Alexander~M. Rush. 2020.
\newblock \href {https://www.aclweb.org/anthology/2020.emnlp-demos.6}
  {Transformers: State-of-the-art natural language processing}.
\newblock In \emph{Proceedings of the 2020 Conference on Empirical Methods in
  Natural Language Processing: System Demonstrations}, pages 38--45, Online.
  Association for Computational Linguistics.

\bibitem[{Xu et~al.(2015)Xu, Ba, Kiros, Cho, Courville, Salakhudinov, Zemel,
  and Bengio}]{pmlr-v37-xuc15}
Kelvin Xu, Jimmy Ba, Ryan Kiros, Kyunghyun Cho, Aaron Courville, Ruslan
  Salakhudinov, Rich Zemel, and Yoshua Bengio. 2015.
\newblock \href {http://proceedings.mlr.press/v37/xuc15.html} {Show, attend and
  tell: Neural image caption generation with visual attention}.
\newblock In \emph{Proceedings of the 32nd International Conference on Machine
  Learning}, volume~37 of \emph{Proceedings of Machine Learning Research},
  pages 2048--2057, Lille, France. PMLR.

\bibitem[{Yang et~al.(2019)Yang, Dai, Yang, Carbonell, Salakhutdinov, and
  Le}]{yang2019xlnet}
Zhilin Yang, Zihang Dai, Yiming Yang, Jaime Carbonell, Ruslan Salakhutdinov,
  and Quoc~V Le. 2019.
\newblock Xlnet: Generalized autoregressive pretraining for language
  understanding.
\newblock \emph{arXiv preprint arXiv:1906.08237}.

\bibitem[{Yao et~al.(2019)Yao, Peng, Weischedel, Knight, Zhao, and
  Yan}]{yao2019plan}
Lili Yao, Nanyun Peng, Ralph Weischedel, Kevin Knight, Dongyan Zhao, and Rui
  Yan. 2019.
\newblock Plan-and-write: Towards better automatic storytelling.
\newblock In \emph{Proceedings of the AAAI Conference on Artificial
  Intelligence}, volume~33, pages 7378--7385.

\bibitem[{Zhang et~al.(2020)Zhang, Wang, Li, Gan, Brockett, and
  Dolan}]{zhang2020pointer}
Yizhe Zhang, Guoyin Wang, Chunyuan Li, Zhe Gan, Chris Brockett, and Bill Dolan.
  2020.
\newblock Pointer: Constrained text generation via insertion-based generative
  pre-training.
\newblock \emph{arXiv preprint arXiv:2005.00558}.

\end{thebibliography}
\bibliographystyle{acl_natbib}

\section*{Checklist}


\begin{enumerate}

\item For all authors...
\begin{enumerate}
  \item Do the main claims made in the abstract and introduction accurately reflect the paper's contributions and scope?
    \answerYes{}
  \item Did you describe the limitations of your work?
    \answerYes{}
  \item Did you discuss any potential negative societal impacts of your work?
    \answerNo{}
  \item Have you read the ethics review guidelines and ensured that your paper conforms to them?
    \answerYes{}
\end{enumerate}

\item If you are including theoretical results...
\begin{enumerate}
  \item Did you state the full set of assumptions of all theoretical results?
    \answerYes{}
        \item Did you include complete proofs of all theoretical results?
    \answerYes{Most theoretical results are self-explanatory.}
\end{enumerate}

\item If you ran experiments...
\begin{enumerate}
  \item Did you include the code, data, and instructions needed to reproduce the main experimental results (either in the supplemental material or as a URL)?
    \answerNo{They will be released upon camera ready.}
  \item Did you specify all the training details (e.g., data splits, hyperparameters, how they were chosen)?
    \answerYes{}
        \item Did you report error bars (e.g., with respect to the random seed after running experiments multiple times)?
    \answerNo{}
        \item Did you include the total amount of compute and the type of resources used (e.g., type of GPUs, internal cluster, or cloud provider)?
    \answerYes{}
\end{enumerate}

\item If you are using existing assets (e.g., code, data, models) or curating/releasing new assets...
\begin{enumerate}
  \item If your work uses existing assets, did you cite the creators?
    \answerYes{}
  \item Did you mention the license of the assets?
    \answerYes{}
  \item Did you include any new assets either in the supplemental material or as a URL?
    \answerYes{}
  \item Did you discuss whether and how consent was obtained from people whose data you're using/curating?
    \answerYes{}
  \item Did you discuss whether the data you are using/curating contains personally identifiable information or offensive content?
    \answerNo{}
\end{enumerate}

\item If you used crowdsourcing or conducted research with human subjects...
\begin{enumerate}
  \item Did you include the full text of instructions given to participants and screenshots, if applicable?
    \answerNA{}
  \item Did you describe any potential participant risks, with links to Institutional Review Board (IRB) approvals, if applicable?
    \answerNA{}
  \item Did you include the estimated hourly wage paid to participants and the total amount spent on participant compensation?
    \answerNA{}
\end{enumerate}

\end{enumerate}


\appendix
\clearpage
\section{Appendix}
\label{sec:appendix}
\subsection{General Setup of Experiments}
For all language generation tasks, based on whether the data is pre-processed upon release or not, either a BPE-based or a classical tokenizer is applied. Each of the evaluated transformer models, if not otherwise stated, is implemented as a \emph{base}-sized transformer model, which has 12-layers with 12 attention head and 768 hidden dimensions. The batch size is set to 256. In cases where the model size exceeds the device capacity, the cumulative gradient trick is applied to support an equivalent optimization effect. The learning rate is selected from \{5e-5, 1e-4, 2e-4\}. The dropout rate is selected from \{0.1, 0.2\}. The weight decay rate is set to 0.02. All the models are trained with 1000 warm-up iterations and a maximum of 200000 iterations of training. A linear-decay learning rate scheduler is applied for fine-grained training of the model. If presented with a development set, the training will be early-stopped when the model stops improving for 5 consecutive epochs. For \model-Dinic, unless otherwise stated, the parallelization fine-tuning lasts for 5 epochs typically.

The setups of MT tasks follow previous works, including, but not limited to InsT/LevT and InDIGO \citep{stern2019insertion,gu2019levenshtein,gu2019insertion}.

Models are implemented based on libraries like PyTorch\citep{paszke2019pytorch} (for the general construction of deep learning models, License: BSD) and huggingface's Transformers\citep{wolf-etal-2020-transformers} (for specific implementation of transformer-based models, License: Apache-2.0). Some of our baselines also used NVIDIA APEX for efficient mix-precision training \footnote{https://github.com/NVIDIA/apex}. We thank them for their contributions to the community. With the best of our effort we use datasets that contain either user-identity agnostic contents or those that could contain publicly reported information about famous, well-known individuals (for the News dataset).

All results that could contain randomness, if not otherwise stated, is collected from averaging the results of 3 individual runs of sampling. The variances of the results won't affect/reverse the conclusion.

\subsection{Extended Study and Qualitative Examples}
Due to the page limit, we could only manage to include some qualitative examples here in the appendix. Table~\ref{appentab:ablation_quan} contains some basic quantitative results for the ablated variants of InsNet. Table~\ref{appentab:ablation} contains a few failure cases of the ablated variants of InsNet. Table~\ref{appentab:traj} contains some randomly sampled trajectories on the News dataset, showing how the context is gradually enriched by InsNet in sequentially-decoded insertion-based text generation.
\begin{wraptable}{r}{.5\textwidth}
\centering
\small
  \renewcommand{\tabcolsep}{.5mm}
  \caption{Ablation study of the position encoding and slot representation.}
  \label{appentab:ablation_quan}
  \resizebox{\linewidth}{!}{
  \begin{tabular}{l@{ }cccc@{}ccccc@{}}
    \toprule
     & InsNet & InsNet-truncated & InsNet-noglobal \\
  \midrule 
  Terminating NLL $\downarrow$ & \textbf{2.17} & 2.97 & 2.56 \\
  \midrule
  BLEU-2/4 $\uparrow$ & \textbf{19.17/5.69} & 16.13/3.89 & 17.83/4.81 \\
  \bottomrule
  \end{tabular}
  }
\end{wraptable}
        \begin{table}[t]
            \centering
            \setlength{\abovecaptionskip}{10pt}
            \setlength{\belowcaptionskip}{0pt}
            \caption{Ablated versions of the model fails to terminate the sequence at reasonable timings, leading to wordy/unreadable outputs in some cases. This is even more severe for samples from earlier checkpoints during the training.}
            \label{appentab:ablation}
            \begin{tabularx}{\textwidth}{lXX}
            \hline
                InsNet & intriguing meals and traditional . fun atmosphere . great prices . great service . \\
                \hline
                InsNet-truncated & ... very intriguing . great service , great staff , and the whole meals are amazing . i love the place . love the food here here . ... (\textbf{non-terminating}) \\
                \hline
                InsNet-noglobal & very intriguing meals . great atmosphere . and service was great . food was amazing . service was delicious . traditional italian cuisine , very fun . atmosphere is great . prices are good and very reasonable . ! . \\
            \end{tabularx}
            \label{tab:case1}
            \begin{tabularx}{\textwidth}{lXX}
            \hline \hline
                InsNet & i have been here several times . great atmosphere and the service is always great . love the food , and the best steak salad . it is fantastic ! \\
                \hline
                InsNet-truncated & i ve been coming here multiple times in years now . great food , good food and delicious . also , the service . great atmosphere and atmosphere . service and food was great . great experience here . love food and salad , and the staff is very friendly ! . ! . \\
                \hline
                InsNet-noglobal & ... everything is very delicious . the whole menu was awesome , especially . we could have ever eaten since we found this place . we have had everything on the food , kids menu , , and large portions and food , great service . great place is awesome . great service , clean , fresh , tasty , tasty , hot , , , good food , service , ... (\textbf{non-terminating}) \\
            \end{tabularx}
            \label{tab:case2}
            \begin{tabularx}{\textwidth}{lXX}
            \hline \hline
                InsNet & this place has been a long time to take out of my favorite spot . service is friendly and fast , the food here are excellent . fantastic sandwiches . this is my favorite spot.  \\
                \hline
                InsNet-truncated & ...  we love it here , and it s always so delicious . the service is amazing . the owner owner is always nice and has been super friendly service , and one of my favorite lunch spot . the food is always fantastic . oh and delicious . and there is the best value . the food is so fresh and the sandwiches are great . is my favorite . , it is fantastic . i .. ... (\textbf{non-terminating}) \\
                \hline
                InsNet-noglobal & delicious food and service . great customer service and food . good food , clean atmosphere . nice staff . good breakfast . great service , fresh food and atmosphere . will come here everyday ! . \\
            \hline
            \end{tabularx}
            \label{tab:case3}
        \end{table}
        
        \begin{table}[t]
            \small
            \centering
            \setlength{\abovecaptionskip}{10pt}
            \setlength{\belowcaptionskip}{0pt}
            \caption{Some generation trajectories collected by randomly sampling from InsNet.}
            \label{appentab:traj}
            \begin{tabularx}{\textwidth}{lXX}
            \hline\hline
                Keywords  &\bb{makes} \bb{election} \bb{lose} \bb{lost} \\
            \hline
                InsNet-News&  \bb{makes} \bb{election} \bb{lose} \bb{lost} \\
                & trump \bb{makes} \bb{election} \bb{lose} \bb{lost} \\
                & trump \bb{makes} the \bb{election} \bb{lose} \bb{lost} \\
                & if trump \bb{makes} the \bb{election} \bb{lose} \bb{lost} \\
                & if trump \bb{makes} the \bb{election} \bb{lose} , \bb{lost} \\
                & if trump \bb{makes} the \bb{election} \bb{lose} , \bb{lost} the \\
                & if trump \bb{makes} the \bb{election} \bb{lose} , \bb{lost} the election \\
                & if trump \bb{makes} the \bb{election} \bb{lose} , \bb{lost} the election . \\
                & if trump \bb{makes} the \bb{election} would \bb{lose} , \bb{lost} the election . \\
                & if trump \bb{makes} the \bb{election} would \bb{lose} , he \bb{lost} the election . \\
                & if trump \bb{makes} the \bb{election} would \bb{lose} election , he \bb{lost} the election . \\
                & " if trump \bb{makes} the \bb{election} would \bb{lose} election , he \bb{lost} the election . \\
                & " if trump \bb{makes} the \bb{election} would \bb{lose} election , he \bb{lost} the election . " \\
                & " if trump \bb{makes} the \bb{election} would \bb{lose} election , he once \bb{lost} the election . " \\
                & " if trump \bb{makes} the \bb{election} would not \bb{lose} election , he once \bb{lost} the election . " \\
                & " if trump \bb{makes} the \bb{election} would not \bb{lose} election again, he once \bb{lost} the election . " \\
                & " if trump \bb{makes} the \bb{election} would not \bb{lose} election again, because he once \bb{lost} the election . " \\
                & " now if trump \bb{makes} the \bb{election} would not \bb{lose} election again, because he once \bb{lost} the election . " \\
                & " now if trump \bb{makes} the \bb{election} would not \bb{lose} this election again, because he once \bb{lost} the election . " \\
                & , " now if trump \bb{makes} the \bb{election} would not \bb{lose} this election again, because he once \bb{lost} the election . " \\
                & the , " now if trump \bb{makes} the \bb{election} would not \bb{lose} this election again, because he once \bb{lost} the election . " \\
                & the , " now if trump \bb{makes} the \bb{election} , would not \bb{lose} this election again, because he once \bb{lost} the election . " \\
                & on the , " now if trump \bb{makes} the \bb{election} , would not \bb{lose} this election again, because he once \bb{lost} the election . " \\
                & on the , " now if trump \bb{makes} the \bb{election} , he would not \bb{lose} this election again, because he once \bb{lost} the election . " \\
                & he on the , " now if trump \bb{makes} the \bb{election} , he would not \bb{lose} this election again, because he once \bb{lost} the election . " \\
                & he said on the , " now if trump \bb{makes} the \bb{election} , he would not \bb{lose} this election again, because he once \bb{lost} the election . " \\
                & he said on the rally , " now if trump \bb{makes} the \bb{election} , he would not \bb{lose} this election again, because he once \bb{lost} the election . " \\
                \hline
                Golden Ref & if he \bb{makes} it to the next \bb{election} he ' ll \bb{lose} , but the other three would have \bb{lost} just as much . \\
                \hline
            \end{tabularx}

        \end{table}
        
        \begin{table}[t]
            \small
            \centering
            \setlength{\abovecaptionskip}{10pt}
            \setlength{\belowcaptionskip}{0pt}
            \begin{tabularx}{\textwidth}{lXX}
            \hline\hline
                Keywords  &\bb{thought} \bb{good} \bb{run} \bb{bring} \\
                \hline
                InsNet-News& \bb{thought} \bb{good} \bb{run} \bb{bring}\\ 
                & \bb{thought} \bb{good} \bb{run} \bb{bring} .\\ 
                & \bb{thought} \bb{good} \bb{run} to \bb{bring} .\\ 
                & i \bb{thought} \bb{good} \bb{run} to \bb{bring} .\\ 
                & i \bb{thought} \bb{good} \bb{run} to \bb{bring} him .\\ 
                & so i \bb{thought} \bb{good} \bb{run} to \bb{bring} him .\\ 
                & so i \bb{thought} would \bb{good} \bb{run} to \bb{bring} him .\\ 
                & so i \bb{thought} it would \bb{good} \bb{run} to \bb{bring} him .\\ 
                & he so i \bb{thought} it would \bb{good} \bb{run} to \bb{bring} him .\\ 
                & he had so i \bb{thought} it would \bb{good} \bb{run} to \bb{bring} him .\\ 
                & he had a so i \bb{thought} it would \bb{good} \bb{run} to \bb{bring} him .\\ 
                & he had a so i \bb{thought} it would be \bb{good} \bb{run} to \bb{bring} him .\\ 
                & he had a , so i \bb{thought} it would be \bb{good} \bb{run} to \bb{bring} him .\\ 
                & he had a , so i \bb{thought} it would be \bb{good} \bb{run} to \bb{bring} to him .\\ 
                & but he had a , so i \bb{thought} it would be \bb{good} \bb{run} to \bb{bring} to him .\\ 
                & but he had a great , so i \bb{thought} it would be \bb{good} \bb{run} to \bb{bring} to him .\\ 
                & but he had a great job , so i \bb{thought} it would be \bb{good} \bb{run} to \bb{bring} to him .\\ 
                & but he had a great job , so i \bb{thought} it would be \bb{good} \bb{run} to \bb{bring} his to him .\\ 
                & but he had a great job in , so i \bb{thought} it would be \bb{good} \bb{run} to \bb{bring} his to him .\\ 
                & but he had a great job in , so i \bb{thought} it would be \bb{good} \bb{run} for to \bb{bring} his to him .\\ 
                & but he had a great job in , so i \bb{thought} it would be \bb{good} \bb{run} for me to \bb{bring} his to him .\\ 
                & but he had a great job in , so i \bb{thought} it would be \bb{good} \bb{run} for me to \bb{bring} his son to him .\\ 
                & but then he had a great job in , so i \bb{thought} it would be \bb{good} \bb{run} for me to \bb{bring} his son to him .\\ 
                & but then he had a great job in , so i \bb{thought} it would be a \bb{good} \bb{run} for me to \bb{bring} his son to him .\\ 
                & but then he had a great job in london , so i \bb{thought} it would be a \bb{good} \bb{run} for me to \bb{bring} his son to him . \\
                \hline
                Golden Ref & i \bb{thought} , " i ' ve had a  \bb{good} \bb{run}  and if this is the way i go ,  \bb{bring} it on . \\
                \hline
                
            \end{tabularx}

        \end{table}

        \begin{table}[t]
            \small
            \centering
            \setlength{\abovecaptionskip}{10pt}
            \setlength{\belowcaptionskip}{0pt}
            \begin{tabularx}{\textwidth}{lXX}
            \hline\hline
                Keywords & \bb{played} \bb{team} \bb{years} \bb{completed} \\
                \hline
                InsNet-News& \bb{played} \bb{team} three \bb{years} \bb{completed} . \\
                & has \bb{played} \bb{team} three \bb{years} \bb{completed} . \\
                & has \bb{played} his \bb{team} three \bb{years} \bb{completed} . \\
                & he has \bb{played} his \bb{team} three \bb{years} \bb{completed} . \\
                & but he has \bb{played} his \bb{team} three \bb{years} \bb{completed} . \\
                & but he has \bb{played} in his \bb{team} three \bb{years} \bb{completed} . \\
                & but he has \bb{played} in his \bb{team} three \bb{years} before \bb{completed} . \\
                & but he has \bb{played} in his \bb{team} three \bb{years} before \bb{completed} the . \\
                & but he has \bb{played} in his \bb{team} three \bb{years} before being \bb{completed} the . \\
                & but he has \bb{played} in his \bb{team} for three \bb{years} before being \bb{completed} the . \\
                & but he has \bb{played} in his \bb{team} for three \bb{years} before being \bb{completed} the year . \\
                & the but he has \bb{played} in his \bb{team} for three \bb{years} before being \bb{completed} the year . \\
                & in the but he has \bb{played} in his \bb{team} for three \bb{years} before being \bb{completed} the year . \\
                & he in the but he has \bb{played} in his \bb{team} for three \bb{years} before being \bb{completed} the year . \\
                & he in the season but he has \bb{played} in his \bb{team} for three \bb{years} before being \bb{completed} the year . \\
                & he has in the season but he has \bb{played} in his \bb{team} for three \bb{years} before being \bb{completed} the year . \\
                & he has not in the season but he has \bb{played} in his \bb{team} for three \bb{years} before being \bb{completed} the year . \\
                & he has not in the last season but he has \bb{played} in his \bb{team} for three \bb{years} before being \bb{completed} the year . \\
                & he has not been in the last season but he has \bb{played} in his \bb{team} for three \bb{years} before being \bb{completed} the year . \\
                & he has not been in the last season but he has \bb{played} in his \bb{team} for three \bb{years} before being \bb{completed} the last year . \\
                \hline
                Golden Ref & he has not \bb{played} for tottenham ' s first \bb{team} since and it is now nearly two \bb{years} since he \bb{completed} a full premier league match for the club . \\

                \hline
            \end{tabularx}

        \end{table}
        \begin{table}[t]
            \small
            \centering
            \setlength{\abovecaptionskip}{10pt}
            \setlength{\belowcaptionskip}{0pt}
            \begin{tabularx}{\textwidth}{lXX}
            \hline\hline
                Keywords & \bb{local} \bb{group} \bb{hurt} \bb{rule} \\
                \hline
                InsNet-News& \bb{local} \bb{group} \bb{hurt} \bb{rule} out \\
                & \bb{local} \bb{group} \bb{hurt} \bb{rule} out of \\
                & \bb{local} \bb{group} \bb{hurt} \bb{rule} out of . \\
                & the \bb{local} \bb{group} \bb{hurt} \bb{rule} out of . \\
                & the \bb{local} \bb{group} \bb{hurt} \bb{rule} out of the . \\
                & the \bb{local} \bb{group} \bb{hurt} \bb{rule} out of the of . \\
                & the \bb{local} \bb{group} \bb{hurt} the \bb{rule} out of the of . \\
                & the \bb{local} \bb{group} \bb{hurt} the \bb{rule} out of the of the . \\
                & the \bb{local} \bb{group} \bb{hurt} the \bb{rule} out of the of the year . \\
                & the \bb{local} \bb{group} \bb{hurt} the government \bb{rule} out of the of the year . \\
                & the \bb{local} \bb{group} has \bb{hurt} the government \bb{rule} out of the of the year . \\
                & the \bb{local} \bb{group} has \bb{hurt} the government \bb{rule} out of the of the last year . \\
                & the \bb{local} \bb{group} has \bb{hurt} the government to \bb{rule} out of the of the last year . \\
                & the \bb{local} \bb{group} has \bb{hurt} the government to \bb{rule} out of the rest of the last year . \\
                & the \bb{local} \bb{group} has been \bb{hurt} the government to \bb{rule} out of the rest of the last year . \\
                & the \bb{local} \bb{group} has been \bb{hurt} the government to \bb{rule} out of for the rest of the last year . \\
                & the \bb{local} \bb{group} has been \bb{hurt} by the government to \bb{rule} out of for the rest of the last year . \\
                & the \bb{local} \bb{group} has been \bb{hurt} by the government to \bb{rule} out of support for the rest of the last year . \\
                \hline 
                Golden Ref& \bb{local} media reported the \bb{group} were not looking to \bb{hurt} anybody , but they would not \bb{rule} out violence if police tried to remove them . \\

                \hline
            \end{tabularx}

        \end{table}
        \begin{table}[t]
            \small
            \centering
            \setlength{\abovecaptionskip}{10pt}
            \setlength{\belowcaptionskip}{0pt}
            \begin{tabularx}{\textwidth}{lXX}
            \hline\hline
                Keywords & \bb{day} \bb{decided} \bb{started} \bb{focus} \\
                \hline
                InsNet-News& \bb{day} \bb{decided} \bb{started} \bb{focus} on \\
                & \bb{day} \bb{decided} \bb{started} \bb{focus} on . \\
                & the \bb{day} \bb{decided} \bb{started} \bb{focus} on . \\
                & the \bb{day} \bb{decided} to \bb{started} \bb{focus} on . \\
                & the \bb{day} , \bb{decided} to \bb{started} \bb{focus} on . \\
                & the \bb{day} , he \bb{decided} to \bb{started} \bb{focus} on . \\
                & the \bb{day} , he \bb{decided} to get \bb{started} \bb{focus} on . \\
                & on the \bb{day} , he \bb{decided} to get \bb{started} \bb{focus} on . \\
                & on the \bb{day} , he \bb{decided} to get \bb{started} \bb{focus} on the . \\
                & on the \bb{day} , he \bb{decided} to get \bb{started} \bb{focus} on the court . \\
                & but on the \bb{day} , he \bb{decided} to get \bb{started} \bb{focus} on the court . \\
                & but on the \bb{day} , he \bb{decided} to get \bb{started} to \bb{focus} on the court . \\
                & but on the next \bb{day} , he \bb{decided} to get \bb{started} to \bb{focus} on the court . \\
                & but , on the next \bb{day} , he \bb{decided} to get \bb{started} to \bb{focus} on the court . \\
                & but , on the next \bb{day} , he \bb{decided} to get \bb{started} to \bb{focus} on the court for . \\
                & but , on the next \bb{day} , he \bb{decided} to get \bb{started} to \bb{focus} on the court for the . \\
                & but , on the next \bb{day} , he \bb{decided} to get \bb{started} to \bb{focus} on the court for the time . \\
                & but , on the next \bb{day} , he \bb{decided} to get \bb{started} to \bb{focus} on the court for the first time . \\
                \hline
                Golden Ref & as the \bb{day} sort of went on and i \bb{decided} to play , i \bb{started} to \bb{focus} a little bit better . \\

                \hline
            \end{tabularx}

        \end{table}
        \begin{table}[t]
            \small
            \centering
            \setlength{\abovecaptionskip}{10pt}
            \setlength{\belowcaptionskip}{0pt}
            \begin{tabularx}{\textwidth}{lXX}
            \hline\hline
                Keywords & \bb{continue} \bb{security} \bb{people} \bb{privacy} \\
                \hline
                InsNet-News& \bb{continue} \bb{security} \bb{people} \bb{privacy} and \\
                & \bb{continue} \bb{security} \bb{people} \bb{privacy} and . \\
                & we \bb{continue} \bb{security} \bb{people} \bb{privacy} and . \\
                & we \bb{continue} \bb{security} of \bb{people} \bb{privacy} and . \\
                & we \bb{continue} \bb{security} of \bb{people} \bb{privacy} and our . \\
                & we \bb{continue} the \bb{security} of \bb{people} \bb{privacy} and our . \\
                & we \bb{continue} the \bb{security} of \bb{people} and \bb{privacy} and our . \\
                & we \bb{continue} the \bb{security} of \bb{people} and \bb{privacy} and our country . \\
                & we \bb{continue} the \bb{security} of \bb{people} and \bb{privacy} and of our country . \\
                & we \bb{continue} the \bb{security} of our \bb{people} and \bb{privacy} and of our country . \\
                & we \bb{continue} to the \bb{security} of our \bb{people} and \bb{privacy} and of our country . \\
                & as we \bb{continue} to the \bb{security} of our \bb{people} and \bb{privacy} and of our country . \\
                & as we \bb{continue} to the \bb{security} of our \bb{people} and \bb{privacy} and safety of our country . \\
                & as we \bb{continue} to protect the \bb{security} of our \bb{people} and \bb{privacy} and safety of our country . \\
                & , as we \bb{continue} to protect the \bb{security} of our \bb{people} and \bb{privacy} and safety of our country . \\
                & , as we \bb{continue} to protect the \bb{security} of our \bb{people} and \bb{privacy} and safety of our country , . \\
                & this , as we \bb{continue} to protect the \bb{security} of our \bb{people} and \bb{privacy} and safety of our country , . \\
                & we this , as we \bb{continue} to protect the \bb{security} of our \bb{people} and \bb{privacy} and safety of our country , . \\
                & we do this , as we \bb{continue} to protect the \bb{security} of our \bb{people} and \bb{privacy} and safety of our country , . \\
                & " we do this , as we \bb{continue} to protect the \bb{security} of our \bb{people} and \bb{privacy} and safety of our country , . \\
                & " we to do this , as we \bb{continue} to protect the \bb{security} of our \bb{people} and \bb{privacy} and safety of our country , . \\
                & " we need to do this , as we \bb{continue} to protect the \bb{security} of our \bb{people} and \bb{privacy} and safety of our country , . \\
                & " we need to do this , as we \bb{continue} to protect the \bb{security} of our \bb{people} and \bb{privacy} and safety of our country , " . \\
                & " we need to do this , as we \bb{continue} to protect the \bb{security} of our \bb{people} and \bb{privacy} and safety of our country , " said . \\
                & " we need to do this , as we \bb{continue} to protect the \bb{security} of our \bb{people} and \bb{privacy} and safety of our country , " he said . \\
                \hline
                Golden Ref & " we ' re going to have to \bb{continue} to balance our needs for \bb{security} with \bb{people} ' s legitimate concerns about \bb{privacy} , " he said . \\

                \hline
            \end{tabularx}

        \end{table}
        
\end{document}